\pdfoutput=1

\documentclass[runningheads]{llncs}
\usepackage[T1]{fontenc}
\usepackage{graphicx}

\usepackage{amsmath}
\usepackage{amssymb}
\usepackage{fvextra}
\makeatletter
\def\PYG@reset{\let\PYG@it=\relax \let\PYG@bf=\relax%
    \let\PYG@ul=\relax \let\PYG@tc=\relax%
    \let\PYG@bc=\relax \let\PYG@ff=\relax}
\def\PYG@tok#1{\csname PYG@tok@#1\endcsname}
\def\PYG@toks#1+{\ifx\relax#1\empty\else%
    \PYG@tok{#1}\expandafter\PYG@toks\fi}
\def\PYG@do#1{\PYG@bc{\PYG@tc{\PYG@ul{%
    \PYG@it{\PYG@bf{\PYG@ff{#1}}}}}}}
\def\PYG#1#2{\PYG@reset\PYG@toks#1+\relax+\PYG@do{#2}}

\@namedef{PYG@tok@w}{\def\PYG@tc##1{\textcolor[rgb]{0.73,0.73,0.73}{##1}}}
\@namedef{PYG@tok@c}{\let\PYG@it=\textit\def\PYG@tc##1{\textcolor[rgb]{0.24,0.48,0.48}{##1}}}
\@namedef{PYG@tok@cp}{\def\PYG@tc##1{\textcolor[rgb]{0.61,0.40,0.00}{##1}}}
\@namedef{PYG@tok@k}{\let\PYG@bf=\textbf\def\PYG@tc##1{\textcolor[rgb]{0.00,0.50,0.00}{##1}}}
\@namedef{PYG@tok@kp}{\def\PYG@tc##1{\textcolor[rgb]{0.00,0.50,0.00}{##1}}}
\@namedef{PYG@tok@kt}{\def\PYG@tc##1{\textcolor[rgb]{0.69,0.00,0.25}{##1}}}
\@namedef{PYG@tok@o}{\def\PYG@tc##1{\textcolor[rgb]{0.40,0.40,0.40}{##1}}}
\@namedef{PYG@tok@ow}{\let\PYG@bf=\textbf\def\PYG@tc##1{\textcolor[rgb]{0.67,0.13,1.00}{##1}}}
\@namedef{PYG@tok@nb}{\def\PYG@tc##1{\textcolor[rgb]{0.00,0.50,0.00}{##1}}}
\@namedef{PYG@tok@nf}{\def\PYG@tc##1{\textcolor[rgb]{0.00,0.00,1.00}{##1}}}
\@namedef{PYG@tok@nc}{\let\PYG@bf=\textbf\def\PYG@tc##1{\textcolor[rgb]{0.00,0.00,1.00}{##1}}}
\@namedef{PYG@tok@nn}{\let\PYG@bf=\textbf\def\PYG@tc##1{\textcolor[rgb]{0.00,0.00,1.00}{##1}}}
\@namedef{PYG@tok@ne}{\let\PYG@bf=\textbf\def\PYG@tc##1{\textcolor[rgb]{0.80,0.25,0.22}{##1}}}
\@namedef{PYG@tok@nv}{\def\PYG@tc##1{\textcolor[rgb]{0.10,0.09,0.49}{##1}}}
\@namedef{PYG@tok@no}{\def\PYG@tc##1{\textcolor[rgb]{0.53,0.00,0.00}{##1}}}
\@namedef{PYG@tok@nl}{\def\PYG@tc##1{\textcolor[rgb]{0.46,0.46,0.00}{##1}}}
\@namedef{PYG@tok@ni}{\let\PYG@bf=\textbf\def\PYG@tc##1{\textcolor[rgb]{0.44,0.44,0.44}{##1}}}
\@namedef{PYG@tok@na}{\def\PYG@tc##1{\textcolor[rgb]{0.41,0.47,0.13}{##1}}}
\@namedef{PYG@tok@nt}{\let\PYG@bf=\textbf\def\PYG@tc##1{\textcolor[rgb]{0.00,0.50,0.00}{##1}}}
\@namedef{PYG@tok@nd}{\def\PYG@tc##1{\textcolor[rgb]{0.67,0.13,1.00}{##1}}}
\@namedef{PYG@tok@s}{\def\PYG@tc##1{\textcolor[rgb]{0.73,0.13,0.13}{##1}}}
\@namedef{PYG@tok@sd}{\let\PYG@it=\textit\def\PYG@tc##1{\textcolor[rgb]{0.73,0.13,0.13}{##1}}}
\@namedef{PYG@tok@si}{\let\PYG@bf=\textbf\def\PYG@tc##1{\textcolor[rgb]{0.64,0.35,0.47}{##1}}}
\@namedef{PYG@tok@se}{\let\PYG@bf=\textbf\def\PYG@tc##1{\textcolor[rgb]{0.67,0.36,0.12}{##1}}}
\@namedef{PYG@tok@sr}{\def\PYG@tc##1{\textcolor[rgb]{0.64,0.35,0.47}{##1}}}
\@namedef{PYG@tok@ss}{\def\PYG@tc##1{\textcolor[rgb]{0.10,0.09,0.49}{##1}}}
\@namedef{PYG@tok@sx}{\def\PYG@tc##1{\textcolor[rgb]{0.00,0.50,0.00}{##1}}}
\@namedef{PYG@tok@m}{\def\PYG@tc##1{\textcolor[rgb]{0.40,0.40,0.40}{##1}}}
\@namedef{PYG@tok@gh}{\let\PYG@bf=\textbf\def\PYG@tc##1{\textcolor[rgb]{0.00,0.00,0.50}{##1}}}
\@namedef{PYG@tok@gu}{\let\PYG@bf=\textbf\def\PYG@tc##1{\textcolor[rgb]{0.50,0.00,0.50}{##1}}}
\@namedef{PYG@tok@gd}{\def\PYG@tc##1{\textcolor[rgb]{0.63,0.00,0.00}{##1}}}
\@namedef{PYG@tok@gi}{\def\PYG@tc##1{\textcolor[rgb]{0.00,0.52,0.00}{##1}}}
\@namedef{PYG@tok@gr}{\def\PYG@tc##1{\textcolor[rgb]{0.89,0.00,0.00}{##1}}}
\@namedef{PYG@tok@ge}{\let\PYG@it=\textit}
\@namedef{PYG@tok@gs}{\let\PYG@bf=\textbf}
\@namedef{PYG@tok@ges}{\let\PYG@bf=\textbf\let\PYG@it=\textit}
\@namedef{PYG@tok@gp}{\let\PYG@bf=\textbf\def\PYG@tc##1{\textcolor[rgb]{0.00,0.00,0.50}{##1}}}
\@namedef{PYG@tok@go}{\def\PYG@tc##1{\textcolor[rgb]{0.44,0.44,0.44}{##1}}}
\@namedef{PYG@tok@gt}{\def\PYG@tc##1{\textcolor[rgb]{0.00,0.27,0.87}{##1}}}
\@namedef{PYG@tok@err}{\def\PYG@bc##1{{\setlength{\fboxsep}{\string -\fboxrule}\fcolorbox[rgb]{1.00,0.00,0.00}{1,1,1}{\strut ##1}}}}
\@namedef{PYG@tok@kc}{\let\PYG@bf=\textbf\def\PYG@tc##1{\textcolor[rgb]{0.00,0.50,0.00}{##1}}}
\@namedef{PYG@tok@kd}{\let\PYG@bf=\textbf\def\PYG@tc##1{\textcolor[rgb]{0.00,0.50,0.00}{##1}}}
\@namedef{PYG@tok@kn}{\let\PYG@bf=\textbf\def\PYG@tc##1{\textcolor[rgb]{0.00,0.50,0.00}{##1}}}
\@namedef{PYG@tok@kr}{\let\PYG@bf=\textbf\def\PYG@tc##1{\textcolor[rgb]{0.00,0.50,0.00}{##1}}}
\@namedef{PYG@tok@bp}{\def\PYG@tc##1{\textcolor[rgb]{0.00,0.50,0.00}{##1}}}
\@namedef{PYG@tok@fm}{\def\PYG@tc##1{\textcolor[rgb]{0.00,0.00,1.00}{##1}}}
\@namedef{PYG@tok@vc}{\def\PYG@tc##1{\textcolor[rgb]{0.10,0.09,0.49}{##1}}}
\@namedef{PYG@tok@vg}{\def\PYG@tc##1{\textcolor[rgb]{0.10,0.09,0.49}{##1}}}
\@namedef{PYG@tok@vi}{\def\PYG@tc##1{\textcolor[rgb]{0.10,0.09,0.49}{##1}}}
\@namedef{PYG@tok@vm}{\def\PYG@tc##1{\textcolor[rgb]{0.10,0.09,0.49}{##1}}}
\@namedef{PYG@tok@sa}{\def\PYG@tc##1{\textcolor[rgb]{0.73,0.13,0.13}{##1}}}
\@namedef{PYG@tok@sb}{\def\PYG@tc##1{\textcolor[rgb]{0.73,0.13,0.13}{##1}}}
\@namedef{PYG@tok@sc}{\def\PYG@tc##1{\textcolor[rgb]{0.73,0.13,0.13}{##1}}}
\@namedef{PYG@tok@dl}{\def\PYG@tc##1{\textcolor[rgb]{0.73,0.13,0.13}{##1}}}
\@namedef{PYG@tok@s2}{\def\PYG@tc##1{\textcolor[rgb]{0.73,0.13,0.13}{##1}}}
\@namedef{PYG@tok@sh}{\def\PYG@tc##1{\textcolor[rgb]{0.73,0.13,0.13}{##1}}}
\@namedef{PYG@tok@s1}{\def\PYG@tc##1{\textcolor[rgb]{0.73,0.13,0.13}{##1}}}
\@namedef{PYG@tok@mb}{\def\PYG@tc##1{\textcolor[rgb]{0.40,0.40,0.40}{##1}}}
\@namedef{PYG@tok@mf}{\def\PYG@tc##1{\textcolor[rgb]{0.40,0.40,0.40}{##1}}}
\@namedef{PYG@tok@mh}{\def\PYG@tc##1{\textcolor[rgb]{0.40,0.40,0.40}{##1}}}
\@namedef{PYG@tok@mi}{\def\PYG@tc##1{\textcolor[rgb]{0.40,0.40,0.40}{##1}}}
\@namedef{PYG@tok@il}{\def\PYG@tc##1{\textcolor[rgb]{0.40,0.40,0.40}{##1}}}
\@namedef{PYG@tok@mo}{\def\PYG@tc##1{\textcolor[rgb]{0.40,0.40,0.40}{##1}}}
\@namedef{PYG@tok@ch}{\let\PYG@it=\textit\def\PYG@tc##1{\textcolor[rgb]{0.24,0.48,0.48}{##1}}}
\@namedef{PYG@tok@cm}{\let\PYG@it=\textit\def\PYG@tc##1{\textcolor[rgb]{0.24,0.48,0.48}{##1}}}
\@namedef{PYG@tok@cpf}{\let\PYG@it=\textit\def\PYG@tc##1{\textcolor[rgb]{0.24,0.48,0.48}{##1}}}
\@namedef{PYG@tok@c1}{\let\PYG@it=\textit\def\PYG@tc##1{\textcolor[rgb]{0.24,0.48,0.48}{##1}}}
\@namedef{PYG@tok@cs}{\let\PYG@it=\textit\def\PYG@tc##1{\textcolor[rgb]{0.24,0.48,0.48}{##1}}}

\def\PYGZbs{\char`\\}
\def\PYGZus{\char`\_}
\def\PYGZob{\char`\{}
\def\PYGZcb{\char`\}}

\def\PYGZlt{\char`\<}
\def\PYGZgt{\char`\>}

\def\PYGZdl{\char`\$}
\def\PYGZhy{\char`\-}
\def\PYGZsq{\char`\'}
\def\PYGZdq{\char`\"}

\makeatother
\begingroup\catcode`\'=\active\gdef\PYGZsq{'}\endgroup
\def\PYGZhy{\mbox{-}}
\VerbatimPygments{\PYG}{\PYG}
\usepackage{url}
\usepackage{tikz}
\usepackage{xcolor}
\usepackage{etoolbox}

\usepackage{newunicodechar}
\newunicodechar{α}{\ensuremath{\alpha}}
\newunicodechar{β}{\ensuremath{\beta}}
\newunicodechar{φ}{\ensuremath{\varphi}}
\newunicodechar{ℓ}{\ensuremath{\ell}}
\newunicodechar{𝕂}{\ensuremath{\mathbb{K}}}
\newunicodechar{𝑻}{\ensuremath{\boldsymbol{T}}}
\newunicodechar{ᴷ}{\ensuremath{^{\mathrm{K}}}}
\newunicodechar{₁}{\ensuremath{_{1}}}
\newunicodechar{₂}{\ensuremath{_{2}}}
\newunicodechar{₃}{\ensuremath{_{3}}}
\newunicodechar{ᴿ}{\ensuremath{^{\mathrm{R}}}}
\newunicodechar{⁺}{\ensuremath{^{+}}}
\newunicodechar{⊔}{\ensuremath{\sqcup}}

\definecolor{agdablue}{HTML}{5E8CB8}

\newif\ifagdarefs
\agdarefsfalse
\newcommand{\AgdaRefs}[1]{\ifagdarefs{\footnotesize\color{agdablue}(#1)}\else\unskip\fi}

\let\claim\relax

\newwrite\claimfile
\immediate\openout\claimfile=\jobname.claims
\newcommand{\claim}[1]{%
  \ifagdarefs{\footnotesize\color{agdablue}[#1]}\else\unskip\fi
  \immediate\write\claimfile{#1}}

\newcommand{\Agda}[2]{{\footnotesize\color{agdablue}\textsf{#1}%
  \ifblank{#2}{}{\texttt{.}\allowbreak\texttt{#2}}}}

\renewcommand{\texttt}[1]{{\footnotesize\color{agdablue}%
  \ttfamily\hyphenchar\font=`\-\relax #1}}
\newcommand{\out}[1]{\mathrm{out}(#1)}
\newcommand{\outM}[3]{\mathrm{out}_{#1,#2}^*(#3)}

\newcommand{\fdef}{:=}

\newcommand{\BB}{\mathbb{B}}
\newcommand{\NN}{\mathbb{N}}

\usepackage[hidelinks]{hyperref}

\begin{document}
\title{Alpha-Beta Pruning via Monadic Continuations}
\titlerunning{Alpha-beta pruning via monadic continuations}

\author{Mart\'{\i}n H\"otzel Escard\'o\inst{1} \and
Paulo Oliva\inst{2}}
\authorrunning{M. H. Escard\'o and P. Oliva}
\institute{Department of Computer Science\\ University of Birmingham\\
\email{m.escardo@bham.ac.uk}\\ \and
School of Electronic Engineering and Computer Science\\ Queen Mary University of London\\
\email{p.oliva@qmul.ac.uk}}
\index{SurnameAuthor1, FirstnameAuthor1}
\index{SurnameAuthor2, FirstnameAuthor2}
\maketitle              %
\begin{abstract} Alpha-beta pruning is a fundamental technique in game-playing AI, enabling efficient minimax search by avoiding the exploration of branches that cannot affect the optimal decision. We present a monadic formulation of alpha-beta pruning based on the combination of the continuation monad and the reader monad. Unlike previous approaches that encode alpha-beta bounds into the types of available moves, our construction records these bounds within monadic outcomes, leaving the move types unchanged. We further show that the ordinary continuation monad suffices to compute not only optimal outcomes but also optimal plays, by extending the outcome type to include the corresponding sequences of moves. This provides an alternative to generalisations of the continuation monad proposed to address the inefficiencies associated with the selection monad in alpha-beta search. We illustrate our approach with an implementation of Tic-Tac-Toe, observing a 39-fold reduction in the number of plays evaluated, compared with exhaustive search, in computing an optimal play. Finally, we formally establish the correctness of our construction by proving that the monadic alpha-beta search computes the same optimal outcomes and plays as exhaustive minimax search. Our results demonstrate how a foundational algorithm in AI can be expressed and verified using standard monadic abstractions, providing a basis for further development of compositional approaches to adversarial search.
{\def\and{{\textperiodcentered} }%
\vspace{2mm}
\par\noindent\textbf{ACM CCS:}\enspace Game tree search \and Functional languages \and Categorical semantics \and Formal software verification \and Type theory
\vspace{2mm}
}
\end{abstract}

\section{Introduction}

Alpha-beta pruning is one of the foundational search techniques in artificial intelligence, particularly in adversarial game playing. One of the earliest explicit descriptions of the method was given by Hart and Edwards in 1961 in the MIT Artificial Intelligence Project memo \emph{The Tree Prune (TP) Algorithm} \cite{hart_edwards_1961}. The method was subsequently described as the \emph{Alpha-Beta Heuristic}, giving rise to the terminology now generally used. Alpha-beta pruning improves minimax search by maintaining bounds on the values available to the maximising and minimising players and terminating the exploration of branches that cannot affect the final decision. The basic algorithm preserves the result of minimax search while avoiding the
evaluation of positions that have been shown to be irrelevant. Its theoretical foundations were later developed more formally by Knuth and Moore \cite{knuth_moore_1975}, establishing alpha-beta pruning as a central technique in game-tree search.

We have shown in \cite{EO(2009),EO(2010D)} that minimax-type searches can be performed using monadic programming, via the continuation monad for calculating optimal outcomes, and the selection monad for optimal plays and strategies (cf. also \cite{EO(2010B),EO(2011A),EO(2023)}). Since the type of the continuation monad
\[
    K_R X \; \equiv \; (X \to R) \to R    
\]
generalises the types of boolean quantifiers $\exists, \forall \colon (X \to \BB) \to \BB$, we have referred to elements of $K_R X$ as \emph{quantifiers}, and used them to model players in so-called \emph{higher-order games}. In $K_R X$, we think of $X$ as the type of 
\emph{available moves} for the player, and $R$ as the type of \emph{game outcomes}.
\AgdaRefs{\Agda{MonadOnTypes.K}{𝕂}, \texttt{K-definitions.K}.}

In a recent work, Hartmann \cite{Hartmann(2026)} proposed an extension of this construction to cater for alpha-beta type optimisations. This builds on the work of Hartmann et al. on a generalisation of the continuation monad \cite{Hartmann(2025)}. In Hartmann \cite{Hartmann(2026)}, the values of $\alpha$ and $\beta$ are ``encoded'' into the types of moves, so that the extended move type becomes $X \times \NN \times \NN$, where $X$ is the actual type of moves, and $\NN \times \NN$ record the current values of alpha and beta.

In the present paper we propose an alternative approach for the monadic alpha-beta optimisation by combining the continuation monad with the reader monad. Our main contributions are as follows:

\begin{itemize}
    \item[(1)] Instead of considering moves of type $X \times \NN \times \NN$, as in \cite{Hartmann(2026)}, we show (Section \ref{sec-alpha-beta}) that the values of alpha and beta can be ``stored'' on monadic outcomes. Formally, let $T$ be a \emph{monad}. We call quantifiers of type $K_{T R} X$ \emph{monadic quantifiers} or \emph{monadic continuation operators} since the result type $T R$ is itself monadic. We will show that by taking $T$ to be the \emph{reader monad}
    \[
        T X = R \times R \to X
    \]
    we can use the monadic outcomes in $T R$ to keep track of the alpha-beta values.
    \AgdaRefs{\Agda{MonadOnTypes.Reader}{Reader}; \Agda{Games.AlphaBeta.Maximizer}{T}.}
    \item[(2)] In Section \ref{sec-proof} we give a formal proof that the monadic continuations with the reader monad indeed calculate the same optimal outcome as the exhaustive search (Corollary \ref{cor-bounds}).
    \claim{same-optimal-outcome}
    \AgdaRefs{\Agda{Games.AlphaBeta.ViaProductsOfQuantifiers}{corollary}.}
    \item[(3)] Hartmann \cite{Hartmann(2026)} and Hartmann et al. \cite{Hartmann(2025)} correctly point out that when carrying the extra information for alpha and beta down the search tree, the \emph{selection monad} seems to recalculate subgames again and again, as the lazy evaluation that allows for efficient selection search is blocked. They suggest a generalisation of the continuation monad (making it polymorphic on the type of outcomes) that enables one to use it to calculate optimal moves. We show here that the usual continuation monad can already be used for such purposes by simply extending the outcome type $R$ to include the play that leads to the outcome (see Section \ref{sec-optimal-plays}).
    \claim{optimal-plays-via-outcomes}
    \AgdaRefs{\Agda{Games.AlphaBeta.OptimalPlays}{optimal-play-of-extended-game},
      \texttt{αβ-play-is-optimal}.}
    \item[(4)] Finally, in Section \ref{sec-history-dependent} we extend the method and its correctness proof to history-dependent games, in which the moves available at a round depend on the moves played so far \cite{EO(2010D),EO(2023)}. We then apply the alpha-beta pruning search in order to compute the optimal outcome and play for Tic-Tac-Toe, and measure the performance gain of using the pruning mechanism.
\end{itemize}

The definitions, results and examples of Sections \ref{sec-proof} and \ref{sec-optimal-plays}, and the outcomes of the example of Sections \ref{sec-exhaustive} and \ref{sec-alpha-beta}, have been formalized in the proof assistant Agda, in the TypeTopology library, in the module Games.AlphaBeta.index and the modules it imports \cite{EO(2026Agda)}. It is developed in the generality of the history-dependent games of Section \ref{sec-history-dependent-products}.
Its module Games.AlphaBeta.Article follows the paper item by item, and its module Games.AlphaBeta.Summary states Theorem \ref{thm-main} and Corollaries \ref{cor-bounds} and \ref{cor-optimal-plays} together with every definition their statements use, so that the statements can be compared with the paper without reading the proofs.

\section{Exhaustive Search via the Continuation Monad}
\label{sec-exhaustive}

We start with an overview of how the continuation monad can be used to perform an exhaustive search on a game tree, using alternating $\max$ and $\min$ quantifiers (cf. \cite{EO(2009)}). We first define the $\min$ and $\max$ quantifiers, then give an example game tree, and finally compose the (local) quantifiers into a (global) quantifier for the whole game.

\subsection{The \texorpdfstring{$\min$}{min} and \texorpdfstring{$\max$}{max} quantifiers}
\label{sec-min-max-quantifiers}

For the sake of the running example, we take integers as moves, lists of integers as the lists of available moves and as plays, and integers as outcomes:

\begin{Verbatim}[commandchars=\\\{\}]
\PYG{k+kr}{import}\PYG{+w}{ }\PYG{n+nn}{K}

\PYG{k+kr}{type}\PYG{+w}{ }\PYG{k+kt}{Move}\PYG{+w}{ }\PYG{o+ow}{=}\PYG{+w}{ }\PYG{k+kt}{Int}
\PYG{k+kr}{type}\PYG{+w}{ }\PYG{k+kt}{Path}\PYG{+w}{ }\PYG{o+ow}{=}\PYG{+w}{ }\PYG{p}{[}\PYG{k+kt}{Move}\PYG{p}{]}
\PYG{k+kr}{type}\PYG{+w}{ }\PYG{k+kt}{Moves}\PYG{+w}{ }\PYG{o+ow}{=}\PYG{+w}{ }\PYG{p}{[}\PYG{k+kt}{Int}\PYG{p}{]}
\PYG{k+kr}{type}\PYG{+w}{ }\PYG{k+kt}{R}\PYG{+w}{ }\PYG{o+ow}{=}\PYG{+w}{ }\PYG{k+kt}{Int}
\end{Verbatim}

We can then define $\min$ and $\max$ quantifiers (players) over these types as:

\begin{Verbatim}[commandchars=\\\{\}]
\PYG{c+c1}{\PYGZhy{}\PYGZhy{} maximising player quantifier}
\PYG{n+nf}{phiMax}\PYG{+w}{ }\PYG{o+ow}{::}\PYG{+w}{ }\PYG{k+kt}{Moves}\PYG{+w}{ }\PYG{o+ow}{\PYGZhy{}\PYGZgt{}}\PYG{+w}{ }\PYG{k+kt}{K}\PYG{+w}{ }\PYG{k+kt}{R}\PYG{+w}{ }\PYG{k+kt}{Move}
\PYG{n+nf}{phiMax}\PYG{+w}{ }\PYG{n}{xs}\PYG{+w}{ }\PYG{o+ow}{=}\PYG{+w}{ }\PYG{k+kt}{K}\PYG{+w}{ }\PYG{p}{(}\PYG{n+nf}{\PYGZbs{}}\PYG{n}{p}\PYG{+w}{ }\PYG{o+ow}{\PYGZhy{}\PYGZgt{}}\PYG{+w}{ }\PYG{n}{maximum}\PYG{+w}{ }\PYG{o}{\PYGZdl{}}\PYG{+w}{ }\PYG{n}{map}\PYG{+w}{ }\PYG{n}{p}\PYG{+w}{ }\PYG{n}{xs}\PYG{p}{)}

\PYG{c+c1}{\PYGZhy{}\PYGZhy{} minimising player quantifier}
\PYG{n+nf}{phiMin}\PYG{+w}{ }\PYG{o+ow}{::}\PYG{+w}{ }\PYG{k+kt}{Moves}\PYG{+w}{ }\PYG{o+ow}{\PYGZhy{}\PYGZgt{}}\PYG{+w}{ }\PYG{k+kt}{K}\PYG{+w}{ }\PYG{k+kt}{R}\PYG{+w}{ }\PYG{k+kt}{Move}
\PYG{n+nf}{phiMin}\PYG{+w}{ }\PYG{n}{xs}\PYG{+w}{ }\PYG{o+ow}{=}\PYG{+w}{ }\PYG{k+kt}{K}\PYG{+w}{ }\PYG{p}{(}\PYG{n+nf}{\PYGZbs{}}\PYG{n}{p}\PYG{+w}{ }\PYG{o+ow}{\PYGZhy{}\PYGZgt{}}\PYG{+w}{ }\PYG{n}{minimum}\PYG{+w}{ }\PYG{o}{\PYGZdl{}}\PYG{+w}{ }\PYG{n}{map}\PYG{+w}{ }\PYG{n}{p}\PYG{+w}{ }\PYG{n}{xs}\PYG{p}{)}
\end{Verbatim}
\AgdaRefs{\Agda{Games.AlphaBeta.Maximizer}{φmax}; \Agda{Games.AlphaBeta.Minimizer}{φmin};
  \Agda{Games.AlphaBeta.Player}{Node}.}

\subsection{An example game tree}
\label{sec-exhaustive-experiment}

In order to illustrate and test the search, we consider the following simple game tree (Fig.~\ref{fig-tree}), based on the example given in the Wikipedia article on alpha-beta pruning \cite{WikipediaAlphaBeta}:

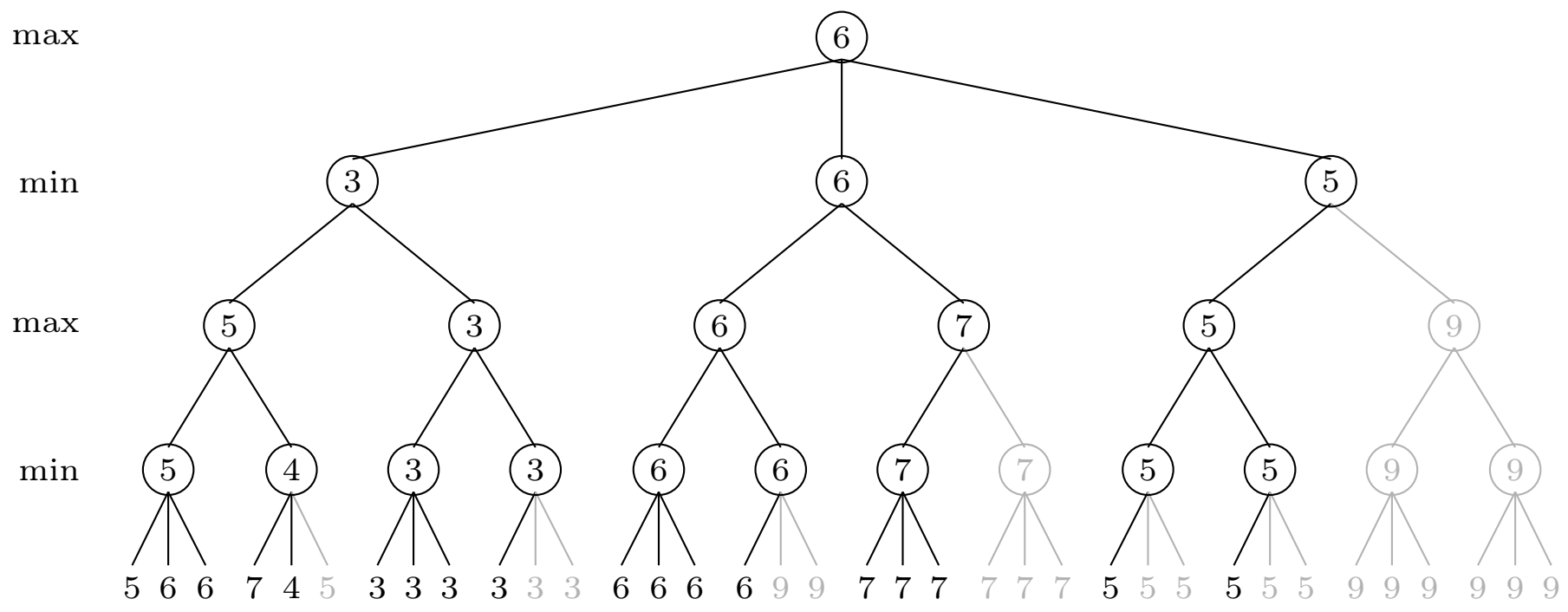
\begin{figure}
\centering
\begin{tikzpicture}[x=0.95cm,y=1cm,font=\scriptsize]
\node[anchor=east] at (-0.30,4.20) {max};
\node[anchor=east] at (-0.30,3.10) {min};
\node[anchor=east] at (-0.30,2.00) {max};
\node[anchor=east] at (-0.30,0.90) {min};
\draw[black] (5.70,4.03) -- (1.77,3.27);
\draw[black] (1.77,2.93) -- (0.78,2.17);
\draw[black] (0.78,1.83) -- (0.29,1.07);
\draw[black] (0.29,0.73) -- (0.00,0.17);
\node[black] at (0.00,0.00) {5};
\draw[black] (0.29,0.73) -- (0.29,0.17);
\node[black] at (0.29,0.00) {6};
\draw[black] (0.29,0.73) -- (0.59,0.17);
\node[black] at (0.59,0.00) {6};
\node[black,draw,circle,inner sep=1pt,minimum size=11pt] at (0.29,0.90) {5};
\draw[black] (0.78,1.83) -- (1.28,1.07);
\draw[black] (1.28,0.73) -- (0.98,0.17);
\node[black] at (0.98,0.00) {7};
\draw[black] (1.28,0.73) -- (1.28,0.17);
\node[black] at (1.28,0.00) {4};
\draw[black!30] (1.28,0.73) -- (1.57,0.17);
\node[black!30] at (1.57,0.00) {5};
\node[black,draw,circle,inner sep=1pt,minimum size=11pt] at (1.28,0.90) {4};
\node[black,draw,circle,inner sep=1pt,minimum size=11pt] at (0.78,2.00) {5};
\draw[black] (1.77,2.93) -- (2.75,2.17);
\draw[black] (2.75,1.83) -- (2.26,1.07);
\draw[black] (2.26,0.73) -- (1.97,0.17);
\node[black] at (1.97,0.00) {3};
\draw[black] (2.26,0.73) -- (2.26,0.17);
\node[black] at (2.26,0.00) {3};
\draw[black] (2.26,0.73) -- (2.55,0.17);
\node[black] at (2.55,0.00) {3};
\node[black,draw,circle,inner sep=1pt,minimum size=11pt] at (2.26,0.90) {3};
\draw[black] (2.75,1.83) -- (3.24,1.07);
\draw[black] (3.24,0.73) -- (2.95,0.17);
\node[black] at (2.95,0.00) {3};
\draw[black!30] (3.24,0.73) -- (3.24,0.17);
\node[black!30] at (3.24,0.00) {3};
\draw[black!30] (3.24,0.73) -- (3.54,0.17);
\node[black!30] at (3.54,0.00) {3};
\node[black,draw,circle,inner sep=1pt,minimum size=11pt] at (3.24,0.90) {3};
\node[black,draw,circle,inner sep=1pt,minimum size=11pt] at (2.75,2.00) {3};
\node[black,draw,circle,inner sep=1pt,minimum size=11pt] at (1.77,3.10) {3};
\draw[black] (5.70,4.03) -- (5.70,3.27);
\draw[black] (5.70,2.93) -- (4.72,2.17);
\draw[black] (4.72,1.83) -- (4.23,1.07);
\draw[black] (4.23,0.73) -- (3.93,0.17);
\node[black] at (3.93,0.00) {6};
\draw[black] (4.23,0.73) -- (4.23,0.17);
\node[black] at (4.23,0.00) {6};
\draw[black] (4.23,0.73) -- (4.52,0.17);
\node[black] at (4.52,0.00) {6};
\node[black,draw,circle,inner sep=1pt,minimum size=11pt] at (4.23,0.90) {6};
\draw[black] (4.72,1.83) -- (5.21,1.07);
\draw[black] (5.21,0.73) -- (4.92,0.17);
\node[black] at (4.92,0.00) {6};
\draw[black!30] (5.21,0.73) -- (5.21,0.17);
\node[black!30] at (5.21,0.00) {9};
\draw[black!30] (5.21,0.73) -- (5.50,0.17);
\node[black!30] at (5.50,0.00) {9};
\node[black,draw,circle,inner sep=1pt,minimum size=11pt] at (5.21,0.90) {6};
\node[black,draw,circle,inner sep=1pt,minimum size=11pt] at (4.72,2.00) {6};
\draw[black] (5.70,2.93) -- (6.68,2.17);
\draw[black] (6.68,1.83) -- (6.19,1.07);
\draw[black] (6.19,0.73) -- (5.90,0.17);
\node[black] at (5.90,0.00) {7};
\draw[black] (6.19,0.73) -- (6.19,0.17);
\node[black] at (6.19,0.00) {7};
\draw[black] (6.19,0.73) -- (6.48,0.17);
\node[black] at (6.48,0.00) {7};
\node[black,draw,circle,inner sep=1pt,minimum size=11pt] at (6.19,0.90) {7};
\draw[black!30] (6.68,1.83) -- (7.17,1.07);
\draw[black!30] (7.17,0.73) -- (6.88,0.17);
\node[black!30] at (6.88,0.00) {7};
\draw[black!30] (7.17,0.73) -- (7.17,0.17);
\node[black!30] at (7.17,0.00) {7};
\draw[black!30] (7.17,0.73) -- (7.47,0.17);
\node[black!30] at (7.47,0.00) {7};
\node[black!30,draw,circle,inner sep=1pt,minimum size=11pt] at (7.17,0.90) {7};
\node[black,draw,circle,inner sep=1pt,minimum size=11pt] at (6.68,2.00) {7};
\node[black,draw,circle,inner sep=1pt,minimum size=11pt] at (5.70,3.10) {6};
\draw[black] (5.70,4.03) -- (9.63,3.27);
\draw[black] (9.63,2.93) -- (8.65,2.17);
\draw[black] (8.65,1.83) -- (8.16,1.07);
\draw[black] (8.16,0.73) -- (7.86,0.17);
\node[black] at (7.86,0.00) {5};
\draw[black!30] (8.16,0.73) -- (8.16,0.17);
\node[black!30] at (8.16,0.00) {5};
\draw[black!30] (8.16,0.73) -- (8.45,0.17);
\node[black!30] at (8.45,0.00) {5};
\node[black,draw,circle,inner sep=1pt,minimum size=11pt] at (8.16,0.90) {5};
\draw[black] (8.65,1.83) -- (9.14,1.07);
\draw[black] (9.14,0.73) -- (8.85,0.17);
\node[black] at (8.85,0.00) {5};
\draw[black!30] (9.14,0.73) -- (9.14,0.17);
\node[black!30] at (9.14,0.00) {5};
\draw[black!30] (9.14,0.73) -- (9.43,0.17);
\node[black!30] at (9.43,0.00) {5};
\node[black,draw,circle,inner sep=1pt,minimum size=11pt] at (9.14,0.90) {5};
\node[black,draw,circle,inner sep=1pt,minimum size=11pt] at (8.65,2.00) {5};
\draw[black!30] (9.63,2.93) -- (10.62,2.17);
\draw[black!30] (10.62,1.83) -- (10.12,1.07);
\draw[black!30] (10.12,0.73) -- (9.83,0.17);
\node[black!30] at (9.83,0.00) {9};
\draw[black!30] (10.12,0.73) -- (10.12,0.17);
\node[black!30] at (10.12,0.00) {9};
\draw[black!30] (10.12,0.73) -- (10.42,0.17);
\node[black!30] at (10.42,0.00) {9};
\node[black!30,draw,circle,inner sep=1pt,minimum size=11pt] at (10.12,0.90) {9};
\draw[black!30] (10.62,1.83) -- (11.11,1.07);
\draw[black!30] (11.11,0.73) -- (10.81,0.17);
\node[black!30] at (10.81,0.00) {9};
\draw[black!30] (11.11,0.73) -- (11.11,0.17);
\node[black!30] at (11.11,0.00) {9};
\draw[black!30] (11.11,0.73) -- (11.40,0.17);
\node[black!30] at (11.40,0.00) {9};
\node[black!30,draw,circle,inner sep=1pt,minimum size=11pt] at (11.11,0.90) {9};
\node[black!30,draw,circle,inner sep=1pt,minimum size=11pt] at (10.62,2.00) {9};
\node[black,draw,circle,inner sep=1pt,minimum size=11pt] at (9.63,3.10) {5};
\node[black,draw,circle,inner sep=1pt,minimum size=11pt] at (5.70,4.20) {6};
\end{tikzpicture}
\caption{The example game tree. Each node shows its optimal outcome. The outcomes in grey are those that the alpha-beta search of Section~\ref{sec-alpha-beta} does not evaluate.}
\label{fig-tree}
\end{figure}

\begin{Verbatim}[commandchars=\\\{\}]
\PYG{k+kr}{import}\PYG{+w}{ }\PYG{n+nn}{Debug.Trace}

\PYG{c+c1}{\PYGZhy{}\PYGZhy{} Example, based on:}
\PYG{c+c1}{\PYGZhy{}\PYGZhy{} https://en.wikipedia.org/w/index.php?oldid=1380302874}

\PYG{c+c1}{\PYGZhy{}\PYGZhy{} Outcome function for the game tree in Example}
\PYG{n+nf}{q\PYGZsq{}}\PYG{+w}{ }\PYG{o+ow}{::}\PYG{+w}{ }\PYG{k+kt}{Path}\PYG{+w}{ }\PYG{o+ow}{\PYGZhy{}\PYGZgt{}}\PYG{+w}{ }\PYG{k+kt}{R}
\PYG{n+nf}{q\PYGZsq{}}\PYG{+w}{ }\PYG{p}{[}\PYG{l+m+mi}{0}\PYG{p}{,}\PYG{l+m+mi}{0}\PYG{p}{,}\PYG{l+m+mi}{0}\PYG{p}{,}\PYG{l+m+mi}{0}\PYG{p}{]}\PYG{+w}{ }\PYG{o+ow}{=}\PYG{+w}{ }\PYG{l+m+mi}{5}
\PYG{n+nf}{q\PYGZsq{}}\PYG{+w}{ }\PYG{p}{[}\PYG{l+m+mi}{0}\PYG{p}{,}\PYG{l+m+mi}{0}\PYG{p}{,}\PYG{l+m+mi}{0}\PYG{p}{,}\PYG{k+kr}{\PYGZus{}}\PYG{p}{]}\PYG{+w}{ }\PYG{o+ow}{=}\PYG{+w}{ }\PYG{l+m+mi}{6}
\PYG{n+nf}{q\PYGZsq{}}\PYG{+w}{ }\PYG{p}{[}\PYG{l+m+mi}{0}\PYG{p}{,}\PYG{l+m+mi}{0}\PYG{p}{,}\PYG{l+m+mi}{1}\PYG{p}{,}\PYG{l+m+mi}{0}\PYG{p}{]}\PYG{+w}{ }\PYG{o+ow}{=}\PYG{+w}{ }\PYG{l+m+mi}{7}
\PYG{n+nf}{q\PYGZsq{}}\PYG{+w}{ }\PYG{p}{[}\PYG{l+m+mi}{0}\PYG{p}{,}\PYG{l+m+mi}{0}\PYG{p}{,}\PYG{l+m+mi}{1}\PYG{p}{,}\PYG{l+m+mi}{1}\PYG{p}{]}\PYG{+w}{ }\PYG{o+ow}{=}\PYG{+w}{ }\PYG{l+m+mi}{4}
\PYG{n+nf}{q\PYGZsq{}}\PYG{+w}{ }\PYG{p}{[}\PYG{l+m+mi}{0}\PYG{p}{,}\PYG{l+m+mi}{0}\PYG{p}{,}\PYG{l+m+mi}{1}\PYG{p}{,}\PYG{l+m+mi}{2}\PYG{p}{]}\PYG{+w}{ }\PYG{o+ow}{=}\PYG{+w}{ }\PYG{l+m+mi}{5}\PYG{+w}{  }\PYG{c+c1}{\PYGZhy{}\PYGZhy{} can be pruned}
\PYG{n+nf}{q\PYGZsq{}}\PYG{+w}{ }\PYG{p}{[}\PYG{l+m+mi}{0}\PYG{p}{,}\PYG{l+m+mi}{1}\PYG{p}{,}\PYG{k+kr}{\PYGZus{}}\PYG{p}{,}\PYG{k+kr}{\PYGZus{}}\PYG{p}{]}\PYG{+w}{ }\PYG{o+ow}{=}\PYG{+w}{ }\PYG{l+m+mi}{3}\PYG{+w}{  }\PYG{c+c1}{\PYGZhy{}\PYGZhy{} can be partially pruned}

\PYG{n+nf}{q\PYGZsq{}}\PYG{+w}{ }\PYG{p}{[}\PYG{l+m+mi}{1}\PYG{p}{,}\PYG{l+m+mi}{0}\PYG{p}{,}\PYG{l+m+mi}{0}\PYG{p}{,}\PYG{k+kr}{\PYGZus{}}\PYG{p}{]}\PYG{+w}{ }\PYG{o+ow}{=}\PYG{+w}{ }\PYG{l+m+mi}{6}
\PYG{n+nf}{q\PYGZsq{}}\PYG{+w}{ }\PYG{p}{[}\PYG{l+m+mi}{1}\PYG{p}{,}\PYG{l+m+mi}{0}\PYG{p}{,}\PYG{l+m+mi}{1}\PYG{p}{,}\PYG{l+m+mi}{0}\PYG{p}{]}\PYG{+w}{ }\PYG{o+ow}{=}\PYG{+w}{ }\PYG{l+m+mi}{6}
\PYG{n+nf}{q\PYGZsq{}}\PYG{+w}{ }\PYG{p}{[}\PYG{l+m+mi}{1}\PYG{p}{,}\PYG{l+m+mi}{0}\PYG{p}{,}\PYG{l+m+mi}{1}\PYG{p}{,}\PYG{k+kr}{\PYGZus{}}\PYG{p}{]}\PYG{+w}{ }\PYG{o+ow}{=}\PYG{+w}{ }\PYG{l+m+mi}{9}\PYG{+w}{  }\PYG{c+c1}{\PYGZhy{}\PYGZhy{} can be pruned}
\PYG{n+nf}{q\PYGZsq{}}\PYG{+w}{ }\PYG{p}{[}\PYG{l+m+mi}{1}\PYG{p}{,}\PYG{l+m+mi}{1}\PYG{p}{,}\PYG{k+kr}{\PYGZus{}}\PYG{p}{,}\PYG{k+kr}{\PYGZus{}}\PYG{p}{]}\PYG{+w}{ }\PYG{o+ow}{=}\PYG{+w}{ }\PYG{l+m+mi}{7}\PYG{+w}{  }\PYG{c+c1}{\PYGZhy{}\PYGZhy{} can be partially pruned}

\PYG{n+nf}{q\PYGZsq{}}\PYG{+w}{ }\PYG{p}{[}\PYG{l+m+mi}{2}\PYG{p}{,}\PYG{l+m+mi}{0}\PYG{p}{,}\PYG{k+kr}{\PYGZus{}}\PYG{p}{,}\PYG{k+kr}{\PYGZus{}}\PYG{p}{]}\PYG{+w}{ }\PYG{o+ow}{=}\PYG{+w}{ }\PYG{l+m+mi}{5}\PYG{+w}{  }\PYG{c+c1}{\PYGZhy{}\PYGZhy{} can be partially pruned}
\PYG{n+nf}{q\PYGZsq{}}\PYG{+w}{ }\PYG{p}{[}\PYG{l+m+mi}{2}\PYG{p}{,}\PYG{k+kr}{\PYGZus{}}\PYG{p}{,}\PYG{k+kr}{\PYGZus{}}\PYG{p}{,}\PYG{k+kr}{\PYGZus{}}\PYG{p}{]}\PYG{+w}{ }\PYG{o+ow}{=}\PYG{+w}{ }\PYG{l+m+mi}{9}\PYG{+w}{  }\PYG{c+c1}{\PYGZhy{}\PYGZhy{} can be pruned}

\PYG{c+c1}{\PYGZhy{}\PYGZhy{} Add tracing to monitor evaluated branches}
\PYG{c+c1}{\PYGZhy{}\PYGZhy{} and order of evaluation}
\PYG{n+nf}{q}\PYG{+w}{ }\PYG{o+ow}{::}\PYG{+w}{ }\PYG{k+kt}{Path}\PYG{+w}{ }\PYG{o+ow}{\PYGZhy{}\PYGZgt{}}\PYG{+w}{ }\PYG{k+kt}{R}
\PYG{n+nf}{q}\PYG{+w}{ }\PYG{n}{xs}\PYG{+w}{ }\PYG{o+ow}{=}\PYG{+w}{ }\PYG{n}{trace}\PYG{+w}{ }\PYG{p}{(}\PYG{n}{show}\PYG{+w}{ }\PYG{n}{xs}\PYG{+w}{ }\PYG{o}{++}\PYG{+w}{ }\PYG{l+s}{\PYGZdq{} \PYGZhy{}\PYGZgt{} \PYGZdq{}}\PYG{+w}{ }\PYG{o}{++}\PYG{+w}{ }\PYG{n}{show}\PYG{+w}{ }\PYG{p}{(}\PYG{n}{q\PYGZsq{}}\PYG{+w}{ }\PYG{n}{xs}\PYG{p}{))}\PYG{+w}{ }\PYG{p}{(}\PYG{n}{q\PYGZsq{}}\PYG{+w}{ }\PYG{n}{xs}\PYG{p}{)}
\end{Verbatim}
\AgdaRefs{\Agda{Games.AlphaBeta.Examples}{wikipedia-tree}, \texttt{wikipedia-nodes},
  \texttt{q}.}

\subsection{Composing quantifiers}
\label{sec-composing-1}

These quantifiers can then be ``sequenced'' to obtain a global quantifier over plays (paths) in the game tree:

\begin{Verbatim}[commandchars=\\\{\}]
\PYG{c+c1}{\PYGZhy{}\PYGZhy{} game quantifiers for the max\PYGZhy{}min experiment game trees}
\PYG{n+nf}{phis}\PYG{+w}{ }\PYG{o+ow}{::}\PYG{+w}{ }\PYG{p}{[}\PYG{k+kt}{K}\PYG{+w}{ }\PYG{k+kt}{R}\PYG{+w}{ }\PYG{k+kt}{Move}\PYG{p}{]}
\PYG{n+nf}{phis}\PYG{+w}{ }\PYG{o+ow}{=}\PYG{+w}{ }\PYG{p}{[}\PYG{n}{phiMax}\PYG{+w}{ }\PYG{p}{[}\PYG{l+m+mi}{0}\PYG{p}{,}\PYG{l+m+mi}{1}\PYG{p}{,}\PYG{l+m+mi}{2}\PYG{p}{],}\PYG{+w}{ }\PYG{n}{phiMin}\PYG{+w}{ }\PYG{p}{[}\PYG{l+m+mi}{0}\PYG{p}{,}\PYG{l+m+mi}{1}\PYG{p}{],}\PYG{+w}{ }\PYG{n}{phiMax}\PYG{+w}{ }\PYG{p}{[}\PYG{l+m+mi}{0}\PYG{p}{,}\PYG{l+m+mi}{1}\PYG{p}{],}\PYG{+w}{ }\PYG{n}{phiMin}\PYG{+w}{ }\PYG{p}{[}\PYG{l+m+mi}{0}\PYG{p}{,}\PYG{l+m+mi}{1}\PYG{p}{,}\PYG{l+m+mi}{2}\PYG{p}{]]}

\PYG{c+c1}{\PYGZhy{}\PYGZhy{} sequencing quantifiers to obtain a \PYGZsq{}global\PYGZsq{} game quantifier}
\PYG{n+nf}{bigPhi}\PYG{+w}{ }\PYG{o+ow}{::}\PYG{+w}{ }\PYG{k+kt}{K}\PYG{+w}{ }\PYG{k+kt}{R}\PYG{+w}{ }\PYG{k+kt}{Path}
\PYG{n+nf}{bigPhi}\PYG{+w}{ }\PYG{o+ow}{=}\PYG{+w}{ }\PYG{n}{sequence}\PYG{+w}{ }\PYG{n}{phis}
\end{Verbatim}
\AgdaRefs{\Agda{Games.AlphaBeta.ViaProductsOfQuantifiers}{quantifier-tree};
  \Agda{Games.FiniteHistoryDependent}{sequenceᴷ}.}

\vspace{3mm}

\begin{remark}[Haskell ``sequence'' and the product of quantifiers] \label{remark-prod-quant} For a monad $M$, the Haskell ``sequence'' operation allows one to combine a sequence of monadic effects into a single effect that ``produces'' a list:
\[
\mathrm{sequence} \; \colon \; (M X)^n \to M X^n.
\]
We have shown in \cite{EO(2009)} that the particular instance of this for the continuation monad (i.e. when $M = K_R$) allows one to compute optimal outcomes in general higher-order games (cf. also \cite{EO(2010D),EO(2011A)}). As mentioned above, we have called elements of the continuation monad $K_R$ \emph{quantifiers}, and the above construction the \emph{product of quantifiers}, using the notation
\[
    \bigotimes \; \colon \; (K_R X)^n \to K_R X^n
\]
which is defined by \AgdaRefs{\Agda{Games.TypeTrees}{path-sequence};
  \Agda{Games.FiniteHistoryDependent}{sequenceᴷ}.}
\[
\begin{array}{rl}
	(\bigotimes [\,])(q) & \fdef q([\,]) \\[2mm]
	(\bigotimes (\phi_0 : \phi))(q) & \fdef \phi_0(\lambda x . (\bigotimes \phi)(\lambda t . q(x \cdot t))),
\end{array}
\]
where $x \cdot t$ adds an element $x$ in front of a sequence $t$. 
\end{remark}

We can then apply the global game quantifier to the (traceable) outcome function to obtain the \emph{optimal outcome} of the game together with the sequence of evaluations of the experimental outcome function $q$:

\begin{Verbatim}[commandchars=\\\{\}]
\PYG{n+nf}{optimal\PYGZus{}outcome}\PYG{+w}{ }\PYG{o+ow}{::}\PYG{+w}{ }\PYG{k+kt}{R}
\PYG{n+nf}{optimal\PYGZus{}outcome}\PYG{+w}{ }\PYG{o+ow}{=}\PYG{+w}{ }\PYG{n}{quantifier}\PYG{+w}{ }\PYG{n}{bigPhi}\PYG{+w}{ }\PYG{n}{q}

\PYG{c+c1}{\PYGZhy{}\PYGZhy{} Test run (all paths are considered):}
\PYG{c+c1}{\PYGZhy{}\PYGZhy{} ghci\PYGZgt{} optimal\PYGZus{}outcome}
\PYG{c+c1}{\PYGZhy{}\PYGZhy{} [0,0,0,0] \PYGZhy{}\PYGZgt{} 5}
\PYG{c+c1}{\PYGZhy{}\PYGZhy{} [0,0,0,1] \PYGZhy{}\PYGZgt{} 6}
\PYG{c+c1}{\PYGZhy{}\PYGZhy{} ...}
\PYG{c+c1}{\PYGZhy{}\PYGZhy{} [2,1,1,1] \PYGZhy{}\PYGZgt{} 9}
\PYG{c+c1}{\PYGZhy{}\PYGZhy{} [2,1,1,2] \PYGZhy{}\PYGZgt{} 9}
\PYG{c+c1}{\PYGZhy{}\PYGZhy{} 6}
\end{Verbatim}

In this case we see that the optimal outcome of the $\max\min$ game was calculated correctly, but all 36 paths were explored.
\claim{exhaustive-outcome}
\AgdaRefs{\Agda{Games.AlphaBeta.ViaProductsOfQuantifiers}{outcome}; \Agda{Games.AlphaBeta.Examples}{wikipedia-outcome}.}

\section{Adding Alpha-Beta Pruning via the Reader Monad}
\label{sec-alpha-beta}

Building on the work of Hartmann \cite{Hartmann(2026)}, we now show how one can implement the alpha-beta optimisation using the standard continuation monad $K_R X \equiv (X \to R) \to R$ by taking the type of outcomes to be $T R$, where $T$ is the reader monad, so that we work with $K_{T R} X$. In \cite{Hartmann(2026)}, instead, the values of alpha and beta were encoded in the type of moves. In our case, the type of moves in the search corresponds to the actual moves of the game, but the monadic outcomes capture the side-effects of updating the values of alpha and beta. We first define the optimised $\min$ and $\max$ quantifiers, then compose them, and finally run the alpha-beta search on the example game tree.

\subsection{The optimised \texorpdfstring{$\min$}{min} and \texorpdfstring{$\max$}{max} quantifiers}
\label{sec-quantifiers-alpha-beta}

Let us start by importing some useful functions, the standard continuation monad (Appendix \ref{app-k}) and the reader monad (from the module Control.Monad.Trans.Reader). The types of moves, paths and outcomes ($R$) are as above.

\begin{Verbatim}[commandchars=\\\{\}]
\PYG{k+kr}{import}\PYG{+w}{ }\PYG{n+nn}{K}
\PYG{k+kr}{import}\PYG{+w}{ }\PYG{n+nn}{Control.Monad.Trans.Reader}
\PYG{k+kr}{import}\PYG{+w}{ }\PYG{n+nn}{Debug.Trace}

\PYG{k+kr}{import}\PYG{+w}{ }\PYG{n+nn}{Utils}

\PYG{c+c1}{\PYGZhy{}\PYGZhy{} Value to be used as \PYGZhy{}inf and +inf}
\PYG{n+nf}{inf}\PYG{+w}{ }\PYG{o+ow}{::}\PYG{+w}{ }\PYG{k+kt}{Int}
\PYG{n+nf}{inf}\PYG{+w}{ }\PYG{o+ow}{=}\PYG{+w}{ }\PYG{l+m+mi}{1000}

\PYG{c+c1}{\PYGZhy{}\PYGZhy{} Reader Monad}
\PYG{k+kr}{type}\PYG{+w}{ }\PYG{k+kt}{T}\PYG{+w}{ }\PYG{o+ow}{=}\PYG{+w}{ }\PYG{k+kt}{Reader}\PYG{+w}{ }\PYG{p}{(}\PYG{k+kt}{Int}\PYG{p}{,}\PYG{+w}{ }\PYG{k+kt}{Int}\PYG{p}{)}
\end{Verbatim}

We can then define quantifiers for the maximising and the minimising player that take the values of $(\alpha, \beta)$ into account. The maximising player stops when the value being considered is greater than or equal to $\beta$, and updates $\alpha$ (highest guaranteed outcome):

\begin{Verbatim}[commandchars=\\\{\}]
\PYG{n+nf}{phiMax}\PYG{+w}{ }\PYG{o+ow}{::}\PYG{+w}{ }\PYG{k+kt}{Moves}\PYG{+w}{ }\PYG{o+ow}{\PYGZhy{}\PYGZgt{}}\PYG{+w}{ }\PYG{p}{(}\PYG{k+kt}{Move}\PYG{+w}{ }\PYG{o+ow}{\PYGZhy{}\PYGZgt{}}\PYG{+w}{ }\PYG{k+kt}{T}\PYG{+w}{ }\PYG{k+kt}{R}\PYG{p}{)}\PYG{+w}{ }\PYG{o+ow}{\PYGZhy{}\PYGZgt{}}\PYG{+w}{ }\PYG{k+kt}{T}\PYG{+w}{ }\PYG{k+kt}{R}
\PYG{n+nf}{phiMax}\PYG{+w}{ }\PYG{n}{xs}\PYG{+w}{ }\PYG{n}{p}\PYG{+w}{ }\PYG{o+ow}{=}\PYG{+w}{ }\PYG{k+kr}{do}
\PYG{+w}{    }\PYG{p}{(}\PYG{k+kr}{\PYGZus{}}\PYG{p}{,}\PYG{+w}{ }\PYG{n}{b}\PYG{p}{)}\PYG{+w}{ }\PYG{o+ow}{\PYGZlt{}\PYGZhy{}}\PYG{+w}{ }\PYG{n}{ask}
\PYG{+w}{    }\PYG{n}{rs}\PYG{+w}{ }\PYG{o+ow}{\PYGZlt{}\PYGZhy{}}\PYG{+w}{ }\PYG{n}{l}\PYG{+w}{ }\PYG{n}{xs}
\PYG{+w}{    }\PYG{n}{return}\PYG{+w}{ }\PYG{o}{\PYGZdl{}}\PYG{+w}{ }\PYG{n}{maximum}\PYG{+w}{ }\PYG{o}{\PYGZdl{}}\PYG{+w}{ }\PYG{n}{takeUntil}\PYG{+w}{ }\PYG{p}{(}\PYG{o}{\PYGZgt{}=}\PYG{+w}{ }\PYG{n}{b}\PYG{p}{)}\PYG{+w}{ }\PYG{n}{rs}
\PYG{+w}{  }\PYG{k+kr}{where}
\PYG{+w}{    }\PYG{n}{l}\PYG{+w}{ }\PYG{o+ow}{::}\PYG{+w}{ }\PYG{k+kt}{Moves}\PYG{+w}{ }\PYG{o+ow}{\PYGZhy{}\PYGZgt{}}\PYG{+w}{ }\PYG{k+kt}{T}\PYG{+w}{ }\PYG{p}{[}\PYG{k+kt}{R}\PYG{p}{]}
\PYG{+w}{    }\PYG{n}{l}\PYG{+w}{ }\PYG{k+kt}{[]}\PYG{+w}{ }\PYG{o+ow}{=}\PYG{+w}{ }\PYG{n}{return}\PYG{+w}{ }\PYG{k+kt}{[]}
\PYG{+w}{    }\PYG{n}{l}\PYG{+w}{ }\PYG{p}{(}\PYG{n}{x}\PYG{k+kt}{:}\PYG{n}{xs}\PYG{p}{)}\PYG{+w}{ }\PYG{o+ow}{=}\PYG{+w}{ }\PYG{k+kr}{do}
\PYG{+w}{      }\PYG{p}{(}\PYG{n}{a}\PYG{p}{,}\PYG{+w}{ }\PYG{n}{b}\PYG{p}{)}\PYG{+w}{ }\PYG{o+ow}{\PYGZlt{}\PYGZhy{}}\PYG{+w}{ }\PYG{n}{ask}
\PYG{+w}{      }\PYG{n}{r}\PYG{+w}{ }\PYG{o+ow}{\PYGZlt{}\PYGZhy{}}\PYG{+w}{ }\PYG{n}{p}\PYG{+w}{ }\PYG{n}{x}
\PYG{+w}{      }\PYG{n}{rs}\PYG{+w}{ }\PYG{o+ow}{\PYGZlt{}\PYGZhy{}}\PYG{+w}{ }\PYG{n}{local}\PYG{+w}{ }\PYG{p}{(}\PYG{n}{const}\PYG{+w}{ }\PYG{p}{(}\PYG{n}{max}\PYG{+w}{ }\PYG{n}{a}\PYG{+w}{ }\PYG{n}{r}\PYG{p}{,}\PYG{+w}{ }\PYG{n}{b}\PYG{p}{))}\PYG{+w}{ }\PYG{p}{(}\PYG{n}{l}\PYG{+w}{ }\PYG{n}{xs}\PYG{p}{)}
\PYG{+w}{      }\PYG{n}{return}\PYG{+w}{ }\PYG{p}{(}\PYG{n}{r}\PYG{+w}{ }\PYG{k+kt}{:}\PYG{+w}{ }\PYG{n}{rs}\PYG{p}{)}
\end{Verbatim}

The minimising player quantifier, on the other hand, stops when the value being considered is less than or equal to $\alpha$, and updates $\beta$ (lowest guaranteed outcome):

\begin{Verbatim}[commandchars=\\\{\}]
\PYG{n+nf}{phiMin}\PYG{+w}{ }\PYG{o+ow}{::}\PYG{+w}{ }\PYG{k+kt}{Moves}\PYG{+w}{ }\PYG{o+ow}{\PYGZhy{}\PYGZgt{}}\PYG{+w}{ }\PYG{p}{(}\PYG{k+kt}{Move}\PYG{+w}{ }\PYG{o+ow}{\PYGZhy{}\PYGZgt{}}\PYG{+w}{ }\PYG{k+kt}{T}\PYG{+w}{ }\PYG{k+kt}{R}\PYG{p}{)}\PYG{+w}{ }\PYG{o+ow}{\PYGZhy{}\PYGZgt{}}\PYG{+w}{ }\PYG{k+kt}{T}\PYG{+w}{ }\PYG{k+kt}{R}
\PYG{n+nf}{phiMin}\PYG{+w}{ }\PYG{n}{xs}\PYG{+w}{ }\PYG{n}{p}\PYG{+w}{ }\PYG{o+ow}{=}\PYG{+w}{ }\PYG{k+kr}{do}
\PYG{+w}{    }\PYG{p}{(}\PYG{n}{a}\PYG{p}{,}\PYG{+w}{ }\PYG{k+kr}{\PYGZus{}}\PYG{p}{)}\PYG{+w}{ }\PYG{o+ow}{\PYGZlt{}\PYGZhy{}}\PYG{+w}{ }\PYG{n}{ask}
\PYG{+w}{    }\PYG{n}{rs}\PYG{+w}{ }\PYG{o+ow}{\PYGZlt{}\PYGZhy{}}\PYG{+w}{ }\PYG{n}{l}\PYG{+w}{ }\PYG{n}{xs}
\PYG{+w}{    }\PYG{n}{return}\PYG{+w}{ }\PYG{o}{\PYGZdl{}}\PYG{+w}{ }\PYG{n}{minimum}\PYG{+w}{ }\PYG{o}{\PYGZdl{}}\PYG{+w}{ }\PYG{n}{takeUntil}\PYG{+w}{ }\PYG{p}{(}\PYG{o}{\PYGZlt{}=}\PYG{+w}{ }\PYG{n}{a}\PYG{p}{)}\PYG{+w}{ }\PYG{n}{rs}
\PYG{+w}{  }\PYG{k+kr}{where}
\PYG{+w}{    }\PYG{n}{l}\PYG{+w}{ }\PYG{o+ow}{::}\PYG{+w}{ }\PYG{k+kt}{Moves}\PYG{+w}{ }\PYG{o+ow}{\PYGZhy{}\PYGZgt{}}\PYG{+w}{ }\PYG{k+kt}{T}\PYG{+w}{ }\PYG{p}{[}\PYG{k+kt}{R}\PYG{p}{]}
\PYG{+w}{    }\PYG{n}{l}\PYG{+w}{ }\PYG{k+kt}{[]}\PYG{+w}{ }\PYG{o+ow}{=}\PYG{+w}{ }\PYG{n}{return}\PYG{+w}{ }\PYG{k+kt}{[]}
\PYG{+w}{    }\PYG{n}{l}\PYG{+w}{ }\PYG{p}{(}\PYG{n}{x}\PYG{k+kt}{:}\PYG{n}{xs}\PYG{p}{)}\PYG{+w}{ }\PYG{o+ow}{=}\PYG{+w}{ }\PYG{k+kr}{do}
\PYG{+w}{      }\PYG{p}{(}\PYG{n}{a}\PYG{p}{,}\PYG{+w}{ }\PYG{n}{b}\PYG{p}{)}\PYG{+w}{ }\PYG{o+ow}{\PYGZlt{}\PYGZhy{}}\PYG{+w}{ }\PYG{n}{ask}
\PYG{+w}{      }\PYG{n}{r}\PYG{+w}{ }\PYG{o+ow}{\PYGZlt{}\PYGZhy{}}\PYG{+w}{ }\PYG{n}{p}\PYG{+w}{ }\PYG{n}{x}
\PYG{+w}{      }\PYG{n}{rs}\PYG{+w}{ }\PYG{o+ow}{\PYGZlt{}\PYGZhy{}}\PYG{+w}{ }\PYG{n}{local}\PYG{+w}{ }\PYG{p}{(}\PYG{n}{const}\PYG{+w}{ }\PYG{p}{(}\PYG{n}{a}\PYG{p}{,}\PYG{+w}{ }\PYG{n}{min}\PYG{+w}{ }\PYG{n}{b}\PYG{+w}{ }\PYG{n}{r}\PYG{p}{))}\PYG{+w}{ }\PYG{p}{(}\PYG{n}{l}\PYG{+w}{ }\PYG{n}{xs}\PYG{p}{)}
\PYG{+w}{      }\PYG{n}{return}\PYG{+w}{ }\PYG{p}{(}\PYG{n}{r}\PYG{+w}{ }\PYG{k+kt}{:}\PYG{+w}{ }\PYG{n}{rs}\PYG{p}{)}
\end{Verbatim}
\AgdaRefs{\Agda{Games.AlphaBeta.Maximizer}{φ*max}; \Agda{Games.AlphaBeta.Minimizer}{φ*min}.}

\subsection{Composing optimised quantifiers}
\label{sec-composing-2}

As in Section \ref{sec-composing-1}, we also use here the monadic ``sequence'' to obtain a single (global) quantifier for the whole game, but now over monadic outcomes $T R$:

\begin{Verbatim}[commandchars=\\\{\}]
\PYG{c+c1}{\PYGZhy{}\PYGZhy{} game quantifiers for max\PYGZhy{}min game trees}
\PYG{n+nf}{phisT}\PYG{+w}{ }\PYG{o+ow}{::}\PYG{+w}{ }\PYG{p}{[}\PYG{k+kt}{K}\PYG{+w}{ }\PYG{p}{(}\PYG{k+kt}{T}\PYG{+w}{ }\PYG{k+kt}{R}\PYG{p}{)}\PYG{+w}{ }\PYG{k+kt}{Move}\PYG{p}{]}
\PYG{n+nf}{phisT}\PYG{+w}{ }\PYG{o+ow}{=}\PYG{+w}{ }\PYG{p}{[}\PYG{k+kt}{K}\PYG{+w}{ }\PYG{p}{(}\PYG{n}{phiMax}\PYG{+w}{ }\PYG{p}{[}\PYG{l+m+mi}{0}\PYG{p}{,}\PYG{l+m+mi}{1}\PYG{p}{,}\PYG{l+m+mi}{2}\PYG{p}{]),}\PYG{+w}{ }\PYG{k+kt}{K}\PYG{+w}{ }\PYG{p}{(}\PYG{n}{phiMin}\PYG{+w}{ }\PYG{p}{[}\PYG{l+m+mi}{0}\PYG{p}{,}\PYG{l+m+mi}{1}\PYG{p}{]),}
\PYG{+w}{         }\PYG{k+kt}{K}\PYG{+w}{ }\PYG{p}{(}\PYG{n}{phiMax}\PYG{+w}{ }\PYG{p}{[}\PYG{l+m+mi}{0}\PYG{p}{,}\PYG{l+m+mi}{1}\PYG{p}{]),}\PYG{+w}{ }\PYG{k+kt}{K}\PYG{+w}{ }\PYG{p}{(}\PYG{n}{phiMin}\PYG{+w}{ }\PYG{p}{[}\PYG{l+m+mi}{0}\PYG{p}{,}\PYG{l+m+mi}{1}\PYG{p}{,}\PYG{l+m+mi}{2}\PYG{p}{])]}

\PYG{c+c1}{\PYGZhy{}\PYGZhy{} sequencing quantifier to obtain \PYGZsq{}global\PYGZsq{} game quantifier}
\PYG{n+nf}{bigPhi}\PYG{+w}{ }\PYG{o+ow}{::}\PYG{+w}{ }\PYG{k+kt}{K}\PYG{+w}{ }\PYG{p}{(}\PYG{k+kt}{T}\PYG{+w}{ }\PYG{k+kt}{R}\PYG{p}{)}\PYG{+w}{ }\PYG{k+kt}{Path}
\PYG{n+nf}{bigPhi}\PYG{+w}{ }\PYG{o+ow}{=}\PYG{+w}{ }\PYG{n}{sequence}\PYG{+w}{ }\PYG{n}{phisT}
\end{Verbatim}
\AgdaRefs{\Agda{Games.AlphaBeta.ViaProductsOfQuantifiers}{quantifier*-tree}.}

\subsection{Alpha-beta search of the example game tree}

We can then test this on the example game of Section \ref{sec-exhaustive-experiment}. We need to make the outcome function $q$ ``monadic'':

\begin{Verbatim}[commandchars=\\\{\}]
\PYG{c+c1}{\PYGZhy{}\PYGZhy{} Make it monadic and add tracing to monitor evaluated branches}
\PYG{c+c1}{\PYGZhy{}\PYGZhy{} and order of evaluation}
\PYG{n+nf}{q}\PYG{+w}{ }\PYG{o+ow}{::}\PYG{+w}{ }\PYG{k+kt}{Path}\PYG{+w}{ }\PYG{o+ow}{\PYGZhy{}\PYGZgt{}}\PYG{+w}{ }\PYG{k+kt}{T}\PYG{+w}{ }\PYG{k+kt}{R}
\PYG{n+nf}{q}\PYG{+w}{ }\PYG{n}{xs}\PYG{+w}{ }\PYG{o+ow}{=}\PYG{+w}{ }\PYG{n}{return}\PYG{+w}{ }\PYG{o}{\PYGZdl{}}\PYG{+w}{ }\PYG{k+kr}{let}\PYG{+w}{ }\PYG{n}{r}\PYG{+w}{ }\PYG{o+ow}{=}\PYG{+w}{ }\PYG{n}{q\PYGZsq{}}\PYG{+w}{ }\PYG{n}{xs}\PYG{+w}{ }\PYG{k+kr}{in}
\PYG{+w}{  }\PYG{n}{trace}\PYG{+w}{ }\PYG{p}{(}\PYG{n}{show}\PYG{+w}{ }\PYG{n}{xs}\PYG{+w}{ }\PYG{o}{++}\PYG{+w}{ }\PYG{l+s}{\PYGZdq{} \PYGZhy{}\PYGZgt{} \PYGZdq{}}\PYG{+w}{ }\PYG{o}{++}\PYG{+w}{ }\PYG{n}{show}\PYG{+w}{ }\PYG{n}{r}\PYG{p}{)}\PYG{+w}{ }\PYG{n}{r}

\PYG{n+nf}{optimal\PYGZus{}outcome}\PYG{+w}{ }\PYG{o+ow}{::}\PYG{+w}{ }\PYG{k+kt}{R}
\PYG{n+nf}{optimal\PYGZus{}outcome}\PYG{+w}{ }\PYG{o+ow}{=}\PYG{+w}{ }\PYG{p}{(}\PYG{n}{runReader}\PYG{+w}{ }\PYG{o}{\PYGZdl{}}\PYG{+w}{ }\PYG{n}{quantifier}\PYG{+w}{ }\PYG{n}{bigPhi}\PYG{+w}{ }\PYG{n}{q}\PYG{p}{)}\PYG{+w}{ }\PYG{p}{(}\PYG{o}{\PYGZhy{}}\PYG{n}{inf}\PYG{p}{,}\PYG{+w}{ }\PYG{n}{inf}\PYG{p}{)}

\PYG{c+c1}{\PYGZhy{}\PYGZhy{} Output:}
\PYG{c+c1}{\PYGZhy{}\PYGZhy{} ghci\PYGZgt{} optimal\PYGZus{}outcome}
\PYG{c+c1}{\PYGZhy{}\PYGZhy{} [0,0,0,0] \PYGZhy{}\PYGZgt{} 5}
\PYG{c+c1}{\PYGZhy{}\PYGZhy{} [0,0,0,1] \PYGZhy{}\PYGZgt{} 6}
\PYG{c+c1}{\PYGZhy{}\PYGZhy{} [0,0,0,2] \PYGZhy{}\PYGZgt{} 6}
\PYG{c+c1}{\PYGZhy{}\PYGZhy{} [0,0,1,0] \PYGZhy{}\PYGZgt{} 7}
\PYG{c+c1}{\PYGZhy{}\PYGZhy{} [0,0,1,1] \PYGZhy{}\PYGZgt{} 4}
\PYG{c+c1}{\PYGZhy{}\PYGZhy{} [0,1,0,0] \PYGZhy{}\PYGZgt{} 3}
\PYG{c+c1}{\PYGZhy{}\PYGZhy{} [0,1,0,1] \PYGZhy{}\PYGZgt{} 3}
\PYG{c+c1}{\PYGZhy{}\PYGZhy{} [0,1,0,2] \PYGZhy{}\PYGZgt{} 3}
\PYG{c+c1}{\PYGZhy{}\PYGZhy{} [0,1,1,0] \PYGZhy{}\PYGZgt{} 3}
\PYG{c+c1}{\PYGZhy{}\PYGZhy{} [1,0,0,0] \PYGZhy{}\PYGZgt{} 6}
\PYG{c+c1}{\PYGZhy{}\PYGZhy{} [1,0,0,1] \PYGZhy{}\PYGZgt{} 6}
\PYG{c+c1}{\PYGZhy{}\PYGZhy{} [1,0,0,2] \PYGZhy{}\PYGZgt{} 6}
\PYG{c+c1}{\PYGZhy{}\PYGZhy{} [1,0,1,0] \PYGZhy{}\PYGZgt{} 6}
\PYG{c+c1}{\PYGZhy{}\PYGZhy{} [1,1,0,0] \PYGZhy{}\PYGZgt{} 7}
\PYG{c+c1}{\PYGZhy{}\PYGZhy{} [1,1,0,1] \PYGZhy{}\PYGZgt{} 7}
\PYG{c+c1}{\PYGZhy{}\PYGZhy{} [1,1,0,2] \PYGZhy{}\PYGZgt{} 7}
\PYG{c+c1}{\PYGZhy{}\PYGZhy{} [2,0,0,0] \PYGZhy{}\PYGZgt{} 5}
\PYG{c+c1}{\PYGZhy{}\PYGZhy{} [2,0,1,0] \PYGZhy{}\PYGZgt{} 5}
\PYG{c+c1}{\PYGZhy{}\PYGZhy{} 6}
\end{Verbatim}

We can see that pruning of the irrelevant branches indeed took place and only the 18 essential plays were considered during the search.
\claim{alpha-beta-outcome}
\AgdaRefs{\Agda{Games.AlphaBeta.ViaProductsOfQuantifiers}{outcome*}; \Agda{Games.AlphaBeta.Examples}{wikipedia-outcome*}.}

\section{Correctness of Alpha-Beta Optimisation}
\label{sec-proof}

We now give a formal proof that the alpha-beta optimisation achieved by the monadic continuations of Section \ref{sec-alpha-beta} indeed (always) computes the same optimal game outcomes as the original (un-optimised) construction. 

We compare outcomes through a map $\mathrm{val} \colon R \to V$ into a type $V$ of values with a decidable strict weak order $<$, that is, a decidable relation that is irreflexive, transitive and cotransitive, where cotransitive means that $a < b$ implies $a < c$ or $c < b$. We write $a \leq b$ for $\neg(b < a)$. Decidable linear orders, such as that of the integers, are examples, and in general a strict weak order is a linear order of classes of tied values, where $a$ and $b$ are tied when neither $a < b$ nor $b < a$. The bounds $\alpha$ and $\beta$ are values. In Sections \ref{sec-exhaustive}--\ref{sec-proof}, $V = R$ and $\mathrm{val}$ is the identity, and in Section \ref{sec-optimal-plays}, $R = V \times \mathrm{Path}$ and $\mathrm{val}$ is the first projection. Comparisons of outcomes, with each other or with values, are comparisons of their values, and $\max(\alpha, r)$ is $\max(\alpha, \mathrm{val}(r))$. The maximum of a non-empty list of outcomes is its first element of largest value, as computed by \Verb[commandchars=\\\{\}]|\PYG{n+nf}{pickMax}| of Appendix \ref{app-utils}, and dually for the minimum. At every position that is not final, the list of available moves is non-empty.
\AgdaRefs{\Agda{OrderedTypes.DecidableStrictWeakOrder}{Decidable-Strict-Weak-Order};
  \Agda{Games.AlphaBeta.Maximizer}{\_⊔ᴿ\_}, \texttt{max⁺}.}

Let $[X]^+$ denote non-empty lists of $X$'s, and let us write 
\[ \phi_{\min}, \phi_{\max} \colon [X]^+ \to (X \to R) \to R \] 
for the standard quantifiers $\min$ and $\max$ (Section \ref{sec-min-max-quantifiers}), and 
\[
\phi_{\min}^*, \phi_{\max}^* \colon [X]^+ \to (X \to T R) \to T R \]
for the monadic versions which perform $\alpha$-$\beta$ pruning where $T$, given by $T X = V \times V \to X$, is the reader monad (Section \ref{sec-quantifiers-alpha-beta}). We will also write $\phi$ or $\phi^*$ as a placeholder for either the $\min$ or the $\max$ version of the corresponding quantifiers.
\AgdaRefs{\Agda{Games.AlphaBeta.ViaProductsOfQuantifiers}{quantifier}, \texttt{quantifier*}.}

Given a game $G = (\phi, q)$, where $\phi$ is a sequence of $n$ quantifiers and $q \colon X^n \to R$ is the outcome function, the \emph{optimal game outcome} can be calculated as (cf. Remark \ref{remark-prod-quant})
\[
\out{G} \fdef \left(\bigotimes \phi \right)(q).
\]
The same can be said for the monadic game $G^* = (\phi^*, \eta \circ q)$, where $\eta(r)(\alpha, \beta) = r$ is the unit of the reader monad. Given concrete values of $(\alpha, \beta)$ we can also obtain a game outcome as 
\[
\outM{\alpha}{\beta}{G} \fdef \left(\bigotimes \phi^* \right)(\eta \circ q)(\alpha, \beta),
\]
\AgdaRefs{\Agda{Games.AlphaBeta.ViaProductsOfQuantifiers}{G}, \texttt{G*}, \texttt{outcome}, \texttt{outcome*}.}
which we will call the \emph{monadic outcome}. We will show that the optimal game outcome indeed coincides with the monadic outcome. Since we will do this by induction on the game tree, we will use the notation $G_s$, given an initial play $s$, for the game $(\phi_s, q_s)$ where $\phi_s$ is the list of quantifiers that remain after playing the moves $s$, and $q_s$, given by $q_s(t) = q(s \cdot t)$, is the outcome function of the subgame defined by $s$. Here $s \cdot t$ is the concatenation of $s$ and $t$ (overloading notation). By the definition of $\bigotimes$, if $\phi$ is the quantifier at $s$, then
\[
\begin{array}{rcl}
\out{G_s} & = & \phi(\lambda x . \out{G_{s\cdot x}}), \\[1mm]
\outM{\alpha}{\beta}{G_s} & = & \phi^*\left(\lambda x, \alpha', \beta' . \outM{\alpha'}{\beta'}{G_{s\cdot x}}\right)(\alpha, \beta).
\end{array}
\]
\AgdaRefs{\Agda{Games.TypeTrees}{subpred}.}
Section \ref{sec-history-dependent} discusses how one can easily extend these definitions and proofs to games in which the quantifier at a position depends on the moves $s$ played so far.

Since we want to prove the correctness of the monadic quantifiers $\phi^*$ of Section \ref{sec-quantifiers-alpha-beta}, let us also define them in standard mathematical notation:

\begin{definition}[$\phi_{\min}^*$ and $\phi_{\max}^*$] \label{def-alpha-beta-quantifiers} Let us first define 
\[ 
\phi_{\max}^* \colon [X]^+ \to (X \to T R) \to T R 
\]
where $[X]^+$ is the type of non-empty lists over $X$, as follows:
\[
\phi_{\max}^*(xs, p)(\alpha, \beta) \fdef \max (\mathrm{takeUntil} \; (\geq \beta) \; \ell(xs) (\alpha, \beta))
\]
where $\ell \colon [X] \to T [R]$
\[
\begin{array}{rcl}
\ell([\,])(\alpha, \beta) & \fdef & [\,] \\[2mm]
\ell(x : xs)(\alpha, \beta) & \fdef & v \colon \ell(xs)(\max(\alpha, v), \beta), \quad \mbox{where $v = p(x)(\alpha, \beta)$.} \end{array}
\]
Here $\mathrm{takeUntil} \; c \; ys$ is the initial segment of $ys$ up to and including the first element that satisfies $c$, and the whole of $ys$ if there is none, so that it is non-empty when $ys$ is.
The definition of $\phi_{\min}^*$ is dual to the above (cf. Haskell code of Section \ref{sec-quantifiers-alpha-beta}). 
\AgdaRefs{\Agda{Games.AlphaBeta.Maximizer}{φ*max}, \texttt{ℓ}; \Agda{MLTT.List-Properties}{take-until};
  \Agda{Games.AlphaBeta.Minimizer}{φ*min}.}
\end{definition}

From now on we focus on $\phi_{\max}^*$. The dual results hold for $\phi_{\min}^*$, since $\phi_{\min}^*$ is $\phi_{\max}^*$ for the reversed order with $\alpha$ and $\beta$ exchanged, which exchanges clauses (i) and (ii) of Theorem \ref{thm-main} below and preserves clause (iii).
\claim{min-dual-to-max}
\AgdaRefs{\Agda{OrderedTypes.DecidableStrictWeakOrder}{opposite}; \Agda{Games.AlphaBeta.Minimizer}{swap},
  \texttt{approximation-to-opposite}, \texttt{approximation-from-opposite}.}

Note that the list $\ell(xs)(\alpha, \beta)$ in the definition of $\phi_{\max}^*$ contains elements of the form $p(x)(\alpha', \beta)$ for different values $\alpha'$. For a list $xs = [x_0, \ldots, x_{n-1}]$, which the search considers in this order, define $\alpha_0, \ldots, \alpha_{n-1}$ by
\[
\alpha_0 := \alpha, \qquad \alpha_{i+1} := \max(\alpha_i, p(x_i)(\alpha_i, \beta)).
\]

\begin{lemma}[Characterisation of $\ell(xs)(\alpha, \beta)$] \label{lemma-alpha-list} Let $p \colon X \to T R$ and $xs = [x_0, \ldots, x_{n-1}] \colon [X]$ and $\alpha, \beta$ be given. Then
\[
\ell(xs)(\alpha, \beta) = [p(x_0)(\alpha_0, \beta), \ldots, p(x_{n-1})(\alpha_{n-1}, \beta)].
\]
\AgdaRefs{\Agda{Games.AlphaBeta.Maximizer}{αs}, \texttt{ℓ-characterization}.}
\end{lemma}

The above lemma shows that the $i$-th member of $\ell(xs)(\alpha, \beta)$ is $p(x_i)(\alpha_i, \beta)$. This will be important for our proof by induction on the game tree. The following lemma is a trivial consequence of the above characterisation of $\ell(xs)(\alpha, \beta)$:

\begin{lemma} \label{main-lemma} Let $p \colon X \to T R$, a non-empty list $xs = [x_0, \ldots, x_{n-1}]$ and $(\alpha, \beta)$ be given. %
\begin{itemize}
    \item[(i)] if $p(x_i)(\alpha, \beta) \leq \alpha$, for all $i < n$, then $\alpha_i = \alpha$ for all $i < n$, and  $\phi_{\max}^*(xs, p)(\alpha, \beta) \leq \alpha$,     
    \item[(ii)] if $p(x_i)(\alpha_i, \beta) \geq \beta$, for some $i < n$, then  $\phi_{\max}^*(xs, p)(\alpha, \beta) \geq \beta$.
\end{itemize}
\AgdaRefs{\Agda{Games.AlphaBeta.Maximizer}{all-αs-are-α}, \texttt{φ*max-is-at-most-α},
  \texttt{φ*max-is-at-least-β}.}
\end{lemma}

As noted above, we can also prove similar (dual) lemmas for $\phi_{\min}^*$ and a sequence $\beta_i$.
\claim{dual-lemmas}
\AgdaRefs{\Agda{Games.AlphaBeta.Minimizer}{Min.ℓ-characterization}, \texttt{Min.all-αs-are-α},
  \texttt{Min.φ*max-is-at-most-α}, \texttt{Min.φ*max-is-at-least-β}.}

\begin{theorem} \label{thm-main} For every partial play $s$ and every pair $(\alpha, \beta)$
\begin{itemize}
    \item[(i)] $\out{G_s} \leq \alpha$ implies $\outM{\alpha}{\beta}{G_s} \leq \alpha$,
    \item[(ii)] $\out{G_s} \geq \beta$ implies $\outM{\alpha}{\beta}{G_s} \geq \beta$,
    \item[(iii)] $\alpha < \out{G_s} < \beta$ implies $\outM{\alpha}{\beta}{G_s} = \out{G_s}$.
\end{itemize}
\AgdaRefs{\Agda{Games.AlphaBeta.ViaProductsOfQuantifiers}{theorem};
  \Agda{Games.AlphaBeta.Maximizer}{\_is-αβ-approximated-by\_}.}
\end{theorem}
\begin{proof} By induction on the game tree. \\[1mm]
(1) If $s$ is a final position, then $\outM{\alpha}{\beta}{G_s} = q(s) = \out{G_s}$, and (i)--(iii) hold trivially. \\[2mm]
(2) Let $s$ be an intermediate position, and suppose w.l.o.g. that $\max$ plays at $s$. Let $xs = [x_0, \ldots, x_{n-1}]$ be the list of moves at $s$, and $p = \lambda x, \alpha', \beta' . \outM{\alpha'}{\beta'}{G_{s\cdot x}}$. We consider three cases: %
\begin{itemize}
    \item[$(i)$] $\out{G_s} \leq \alpha$. This is the simplest case because if  $\out{G_s} \leq \alpha$ then $\out{G_{s\cdot x}} \leq \alpha$, for all moves $x$ (since we are considering the $\max$ player). By induction hypothesis, we have that $\outM{\alpha}{\beta}{G_{s\cdot x}} \leq \alpha$, for all $x$. In this case, in the definition of the recursive function $\ell$, the values of $\alpha$ and $\beta$ are not updated. Hence
    \[
        \begin{array}{lcl}
        \outM{\alpha}{\beta}{G_s}             & = & \phi_{\max}^*(xs, p)(\alpha, \beta) \\[2mm]
            & \stackrel{\textup{L\ref{main-lemma}\,$(i)$}}{\leq} & \alpha.
        \end{array}
    \]       %
    \AgdaRefs{\Agda{Games.AlphaBeta.Maximizer}{part₁}.}
    \item[$(ii)$] $\out{G_s} \geq \beta$. Let $i$ be such that $\out{G_{s\cdot x_i}} \geq \beta$. Since $\max$ updates $\alpha$ but not $\beta$, IH (ii) at $(\alpha_i, \beta)$ gives $\outM{\alpha_i}{\beta}{G_{s\cdot x_i}} \geq \beta$, and hence, by Lemma~\ref{main-lemma}(ii),
    \[
        \begin{array}{lcl}
        \outM{\alpha}{\beta}{G_s}             & = & \phi_{\max}^*(xs, p)(\alpha, \beta) \\[2mm]
            & \stackrel{\textup{L\ref{main-lemma}\,$(ii)$}}{\geq} & \beta.
        \end{array}
    \]       %
    \AgdaRefs{\Agda{Games.AlphaBeta.Maximizer}{part₂}.}
    \item[$(iii)$] $\alpha < \out{G_s} < \beta$. Let $v = \out{G_s}$, and let $k$ be the index at which this maximum is attained first, so that $\out{G_{s\cdot x_k}} = v$. Then $\out{G_{s\cdot x_i}} \leq v < \beta$ for every $i$, and $\out{G_{s\cdot x_i}} < v$ for $i < k$. Each index $i$ falls under one of the following cases:
    \begin{enumerate}
        \item[(a)] $\out{G_{s\cdot x_i}} \leq \alpha_i$, and then, by IH (i), $\outM{\alpha_i}{\beta}{G_{s\cdot x_i}} \leq \alpha_i$;
        \item[(b)] $\alpha_i < \out{G_{s\cdot x_i}}$, and then, since $\out{G_{s\cdot x_i}} < \beta$, IH (iii) gives $\outM{\alpha_i}{\beta}{G_{s\cdot x_i}} = \out{G_{s\cdot x_i}}$.
    \end{enumerate}
    Since $\alpha < v$, induction on $i$ gives $\alpha_i < v$ and $\outM{\alpha_i}{\beta}{G_{s\cdot x_i}} < v$ for every $i < k$. Hence $\alpha_k < v$, so (b) applies to $k$, and
    \[ \outM{\alpha_k}{\beta}{G_{s\cdot x_k}} = \out{G_{s\cdot x_k}} = v. \]
    For $k < i$ we have $\alpha_i = v$, and so $\outM{\alpha_i}{\beta}{G_{s\cdot x_i}} \leq v$, by (a) or (b). Therefore no member of the list is $\geq \beta$, $\mathrm{takeUntil} \; (\geq \beta)$ keeps the whole list, and its maximum is $v$, attained first at the index $k$.
    Hence,
    \[
        \begin{array}{lcl}
        \outM{\alpha}{\beta}{G_s}             & = & \phi_{\max}^*(xs, p)(\alpha, \beta) \\[2mm]
            & = & \max (\mathrm{takeUntil} \; (\geq \beta) \; \ell(xs)(\alpha, \beta)) \\[1mm]
            & \stackrel{\textup{L}\ref{lemma-alpha-list}}{=} & \max(\mathrm{takeUntil} \; (\geq \beta) \; [\ldots, \outM{\alpha_k}{\beta}{G_{s\cdot x_k}}, \ldots]) \\[2mm]
            & = & \outM{\alpha_k}{\beta}{G_{s\cdot x_k}} \\[2mm]
            & = & \out{G_{s\cdot x_k}} \\[2mm]
            & = & \phi_{\max} \left( \lambda x . \out{G_{s\cdot x}} \right) \\[2mm]
            & = & \out{G_s}.
        \end{array}
    \]
    \AgdaRefs{\Agda{Games.AlphaBeta.Maximizer}{part₃}.}
\end{itemize}
\AgdaRefs{\Agda{Games.AlphaBeta.Maximizer}{φmax-is-αβ-approximated-by-φ*max};
  \Agda{Games.AlphaBeta.Minimizer}{φmin-is-αβ-approximated-by-φ*min}.} \qed
\end{proof}

\begin{corollary} \label{cor-bounds}
\begin{itemize}
    \item[(a)] If $\alpha < q(t) < \beta$ for every play $t$, then $\outM{\alpha}{\beta}{G} = \out{G}$.
    \item[(b)] If $\alpha < \outM{\alpha}{\beta}{G} < \beta$, then $\outM{\alpha}{\beta}{G} = \out{G}$.
\end{itemize}
\AgdaRefs{\Agda{Games.AlphaBeta.ViaProductsOfQuantifiers}{corollary}, \texttt{outcome-preserves}, \texttt{corollary'}.}
\end{corollary}

\begin{proof} (a) $\out{G}$ is one of the values $q(t)$, so Theorem \ref{thm-main}(iii) applies. (b) By Theorem \ref{thm-main}(i) and (ii), $\out{G} \leq \alpha$ and $\out{G} \geq \beta$ are impossible, so (iii) applies. \qed
\end{proof}

\section{Computing Optimal Plays}
\label{sec-optimal-plays}

In Section \ref{sec-alpha-beta} we showed how to use the following instance of the continuation monad 
\[ (X \to T R) \to T R, \] 
with $R$ the type of game outcomes and $T$ the reader monad, to simulate the alpha-beta pruning optimisation of the exhaustive search. This allowed us to compute the \emph{optimal outcome} of a game in a more efficient way.

We now show that with a few minor changes we can compute not just the \emph{optimal outcome} but also an \emph{optimal play} that leads to it, while still making use of the alpha-beta optimisation.

Our first change is to extend the outcome type $R$ to include the play that leads to the outcome: 

\begin{Verbatim}[commandchars=\\\{\}]
\PYG{c+c1}{\PYGZhy{}\PYGZhy{} Extended outcome type to include play leading to the outcome}
\PYG{k+kr}{type}\PYG{+w}{ }\PYG{k+kt}{R}\PYG{+w}{ }\PYG{o+ow}{=}\PYG{+w}{ }\PYG{p}{(}\PYG{k+kt}{Int}\PYG{p}{,}\PYG{+w}{ }\PYG{k+kt}{Path}\PYG{p}{)}
\end{Verbatim}

We then need to make four minor changes to the minmax quantifiers as the type $R$ is no longer just an integer, but a pair of an integer and a path. For instance, the amended $\max$ quantifier would become (changes annotated with comments):

\begin{Verbatim}[commandchars=\\\{\}]
\PYG{n+nf}{phiMax}\PYG{+w}{ }\PYG{o+ow}{::}\PYG{+w}{ }\PYG{k+kt}{Moves}\PYG{+w}{ }\PYG{o+ow}{\PYGZhy{}\PYGZgt{}}\PYG{+w}{ }\PYG{k+kt}{K}\PYG{+w}{ }\PYG{p}{(}\PYG{k+kt}{T}\PYG{+w}{ }\PYG{k+kt}{R}\PYG{p}{)}\PYG{+w}{ }\PYG{k+kt}{Move}
\PYG{n+nf}{phiMax}\PYG{+w}{ }\PYG{o+ow}{=}\PYG{+w}{ }\PYG{k+kt}{K}\PYG{+w}{ }\PYG{o}{.}\PYG{+w}{ }\PYG{n}{phiMax\PYGZsq{}}
\PYG{+w}{  }\PYG{k+kr}{where}
\PYG{+w}{    }\PYG{n}{phiMax\PYGZsq{}}\PYG{+w}{ }\PYG{o+ow}{::}\PYG{+w}{ }\PYG{k+kt}{Moves}\PYG{+w}{ }\PYG{o+ow}{\PYGZhy{}\PYGZgt{}}\PYG{+w}{ }\PYG{p}{(}\PYG{k+kt}{Move}\PYG{+w}{ }\PYG{o+ow}{\PYGZhy{}\PYGZgt{}}\PYG{+w}{ }\PYG{k+kt}{T}\PYG{+w}{ }\PYG{k+kt}{R}\PYG{p}{)}\PYG{+w}{ }\PYG{o+ow}{\PYGZhy{}\PYGZgt{}}\PYG{+w}{ }\PYG{k+kt}{T}\PYG{+w}{ }\PYG{k+kt}{R}
\PYG{+w}{    }\PYG{n}{phiMax\PYGZsq{}}\PYG{+w}{ }\PYG{n}{xs}\PYG{+w}{ }\PYG{n}{p}\PYG{+w}{ }\PYG{o+ow}{=}\PYG{+w}{ }\PYG{k+kr}{do}
\PYG{+w}{      }\PYG{p}{(}\PYG{k+kr}{\PYGZus{}}\PYG{p}{,}\PYG{+w}{ }\PYG{n}{b}\PYG{p}{)}\PYG{+w}{ }\PYG{o+ow}{\PYGZlt{}\PYGZhy{}}\PYG{+w}{ }\PYG{n}{ask}
\PYG{+w}{      }\PYG{n}{rs}\PYG{+w}{ }\PYG{o+ow}{\PYGZlt{}\PYGZhy{}}\PYG{+w}{ }\PYG{n}{results}\PYG{+w}{ }\PYG{n}{xs}\PYG{+w}{ }\PYG{n}{p}
\PYG{+w}{      }\PYG{c+c1}{\PYGZhy{}\PYGZhy{} changed (\PYGZgt{}=b) to \PYGZbs{}(r, ms) \PYGZhy{}\PYGZgt{} r \PYGZgt{}= b}
\PYG{+w}{      }\PYG{c+c1}{\PYGZhy{}\PYGZhy{} changed maximum to pickMax}
\PYG{+w}{      }\PYG{n}{return}\PYG{+w}{ }\PYG{o}{\PYGZdl{}}\PYG{+w}{ }\PYG{n}{pickMax}\PYG{+w}{ }\PYG{o}{\PYGZdl{}}\PYG{+w}{ }\PYG{n}{takeUntil}\PYG{+w}{ }\PYG{p}{(}\PYG{n+nf}{\PYGZbs{}}\PYG{p}{(}\PYG{n}{r}\PYG{p}{,}\PYG{+w}{ }\PYG{n}{ms}\PYG{p}{)}\PYG{+w}{ }\PYG{o+ow}{\PYGZhy{}\PYGZgt{}}\PYG{+w}{ }\PYG{n}{r}\PYG{+w}{ }\PYG{o}{\PYGZgt{}=}\PYG{+w}{ }\PYG{n}{b}\PYG{p}{)}\PYG{+w}{ }\PYG{n}{rs}

\PYG{+w}{    }\PYG{n}{results}\PYG{+w}{ }\PYG{o+ow}{::}\PYG{+w}{ }\PYG{k+kt}{Moves}\PYG{+w}{ }\PYG{o+ow}{\PYGZhy{}\PYGZgt{}}\PYG{+w}{ }\PYG{p}{(}\PYG{k+kt}{Move}\PYG{+w}{ }\PYG{o+ow}{\PYGZhy{}\PYGZgt{}}\PYG{+w}{ }\PYG{k+kt}{T}\PYG{+w}{ }\PYG{k+kt}{R}\PYG{p}{)}\PYG{+w}{ }\PYG{o+ow}{\PYGZhy{}\PYGZgt{}}\PYG{+w}{ }\PYG{k+kt}{T}\PYG{+w}{ }\PYG{p}{[}\PYG{k+kt}{R}\PYG{p}{]}
\PYG{+w}{    }\PYG{n}{results}\PYG{+w}{ }\PYG{k+kt}{[]}\PYG{+w}{ }\PYG{k+kr}{\PYGZus{}}\PYG{+w}{ }\PYG{o+ow}{=}\PYG{+w}{ }\PYG{n}{return}\PYG{+w}{ }\PYG{k+kt}{[]}
\PYG{+w}{    }\PYG{n}{results}\PYG{+w}{ }\PYG{p}{(}\PYG{n}{y}\PYG{k+kt}{:}\PYG{n}{ys}\PYG{p}{)}\PYG{+w}{ }\PYG{n}{p}\PYG{+w}{ }\PYG{o+ow}{=}\PYG{+w}{ }\PYG{k+kr}{do}
\PYG{+w}{      }\PYG{p}{(}\PYG{n}{a}\PYG{p}{,}\PYG{+w}{ }\PYG{n}{b}\PYG{p}{)}\PYG{+w}{ }\PYG{o+ow}{\PYGZlt{}\PYGZhy{}}\PYG{+w}{ }\PYG{n}{ask}
\PYG{+w}{      }\PYG{c+c1}{\PYGZhy{}\PYGZhy{} changed r \PYGZlt{}\PYGZhy{} p y to (r, ms) \PYGZlt{}\PYGZhy{} p y}
\PYG{+w}{      }\PYG{p}{(}\PYG{n}{r}\PYG{p}{,}\PYG{+w}{ }\PYG{n}{ms}\PYG{p}{)}\PYG{+w}{ }\PYG{o+ow}{\PYGZlt{}\PYGZhy{}}\PYG{+w}{ }\PYG{n}{p}\PYG{+w}{ }\PYG{n}{y}
\PYG{+w}{      }\PYG{n}{rs}\PYG{+w}{ }\PYG{o+ow}{\PYGZlt{}\PYGZhy{}}\PYG{+w}{ }\PYG{n}{local}\PYG{+w}{ }\PYG{p}{(}\PYG{n}{const}\PYG{+w}{ }\PYG{p}{(}\PYG{n}{max}\PYG{+w}{ }\PYG{n}{a}\PYG{+w}{ }\PYG{n}{r}\PYG{p}{,}\PYG{+w}{ }\PYG{n}{b}\PYG{p}{))}\PYG{+w}{ }\PYG{p}{(}\PYG{n}{results}\PYG{+w}{ }\PYG{n}{ys}\PYG{+w}{ }\PYG{n}{p}\PYG{p}{)}
\PYG{+w}{      }\PYG{c+c1}{\PYGZhy{}\PYGZhy{} changed r to (r, ms)}
\PYG{+w}{      }\PYG{n}{return}\PYG{+w}{ }\PYG{p}{((}\PYG{n}{r}\PYG{p}{,}\PYG{+w}{ }\PYG{n}{ms}\PYG{p}{)}\PYG{+w}{ }\PYG{k+kt}{:}\PYG{+w}{ }\PYG{n}{rs}\PYG{p}{)}
\end{Verbatim}
\AgdaRefs{\Agda{Games.AlphaBeta.Maximizer}{φ*max}, \texttt{max⁺}, with the valuation \Agda{Games.AlphaBeta.OptimalPlays}{value}.}

\vspace{2mm}

The function \Verb[commandchars=\\\{\}]|\PYG{n+nf}{pickMax}| chooses the first of two outcomes with the same value. This matters: a pruned subgame returns an outcome whose value is only a bound, which must not displace an earlier outcome with the same value. The amended min quantifier is obtained in the same way from the \Verb[commandchars=\\\{\}]|\PYG{n+nf}{phiMin}| of Section~\ref{sec-quantifiers-alpha-beta}, with \Verb[commandchars=\\\{\}]|\PYG{n+nf}{pickMin}| in place of \Verb[commandchars=\\\{\}]|\PYG{n+nf}{minimum}|. Since \Verb[commandchars=\\\{\}]|\PYG{n+nf}{phiMax}| and \Verb[commandchars=\\\{\}]|\PYG{n+nf}{phiMin}| now return elements of \Verb[commandchars=\\\{\}]|\PYG{k+kt}{K}|, the list of quantifiers becomes:

\begin{Verbatim}[commandchars=\\\{\}]
\PYG{n+nf}{phisT}\PYG{+w}{ }\PYG{o+ow}{::}\PYG{+w}{ }\PYG{p}{[}\PYG{k+kt}{K}\PYG{+w}{ }\PYG{p}{(}\PYG{k+kt}{T}\PYG{+w}{ }\PYG{k+kt}{R}\PYG{p}{)}\PYG{+w}{ }\PYG{k+kt}{Move}\PYG{p}{]}
\PYG{n+nf}{phisT}\PYG{+w}{ }\PYG{o+ow}{=}\PYG{+w}{ }\PYG{p}{[}\PYG{n}{phiMax}\PYG{+w}{ }\PYG{p}{[}\PYG{l+m+mi}{0}\PYG{p}{,}\PYG{l+m+mi}{1}\PYG{p}{,}\PYG{l+m+mi}{2}\PYG{p}{],}\PYG{+w}{ }\PYG{n}{phiMin}\PYG{+w}{ }\PYG{p}{[}\PYG{l+m+mi}{0}\PYG{p}{,}\PYG{l+m+mi}{1}\PYG{p}{],}\PYG{+w}{ }\PYG{n}{phiMax}\PYG{+w}{ }\PYG{p}{[}\PYG{l+m+mi}{0}\PYG{p}{,}\PYG{l+m+mi}{1}\PYG{p}{],}\PYG{+w}{ }\PYG{n}{phiMin}\PYG{+w}{ }\PYG{p}{[}\PYG{l+m+mi}{0}\PYG{p}{,}\PYG{l+m+mi}{1}\PYG{p}{,}\PYG{l+m+mi}{2}\PYG{p}{]]}
\end{Verbatim}

We then need to modify the outcome function $q$ so that it returns not just the game outcome, but also the play that led to that outcome:

\begin{Verbatim}[commandchars=\\\{\}]
\PYG{c+c1}{\PYGZhy{}\PYGZhy{} Make it monadic and add tracing to monitor evaluated branches}
\PYG{c+c1}{\PYGZhy{}\PYGZhy{} and order of evaluation}
\PYG{n+nf}{q}\PYG{+w}{ }\PYG{o+ow}{::}\PYG{+w}{ }\PYG{k+kt}{Path}\PYG{+w}{ }\PYG{o+ow}{\PYGZhy{}\PYGZgt{}}\PYG{+w}{ }\PYG{k+kt}{T}\PYG{+w}{ }\PYG{k+kt}{R}
\PYG{n+nf}{q}\PYG{+w}{ }\PYG{n}{xs}\PYG{+w}{ }\PYG{o+ow}{=}\PYG{+w}{ }\PYG{n}{return}\PYG{+w}{ }\PYG{o}{\PYGZdl{}}\PYG{+w}{ }\PYG{k+kr}{let}\PYG{+w}{ }\PYG{n}{r}\PYG{+w}{ }\PYG{o+ow}{=}\PYG{+w}{ }\PYG{n}{q\PYGZsq{}}\PYG{+w}{ }\PYG{n}{xs}\PYG{+w}{ }\PYG{k+kr}{in}
\PYG{+w}{         }\PYG{n}{trace}\PYG{+w}{ }\PYG{p}{(}\PYG{n}{show}\PYG{+w}{ }\PYG{n}{xs}\PYG{+w}{ }\PYG{o}{++}\PYG{+w}{ }\PYG{l+s}{\PYGZdq{} \PYGZhy{}\PYGZgt{} \PYGZdq{}}\PYG{+w}{ }\PYG{o}{++}\PYG{+w}{ }\PYG{n}{show}\PYG{+w}{ }\PYG{n}{r}\PYG{p}{)}\PYG{+w}{ }\PYG{p}{(}\PYG{n}{r}\PYG{p}{,}\PYG{+w}{ }\PYG{n}{xs}\PYG{p}{)}
\end{Verbatim}
\AgdaRefs{\Agda{Games.AlphaBeta.OptimalPlays}{q'}.}

\vspace{3mm}

A run of the game quantifier on this outcome function will not just compute the optimal outcome but also an optimal play (while still performing the alpha-beta pruning):
\claim{alpha-beta-optimal-play}
\AgdaRefs{\Agda{Games.AlphaBeta.OptimalPlays}{αβ-outcome-and-play},
  \texttt{αβ-outcome-is-optimal}, \texttt{αβ-play-is-optimal};
  \Agda{Games.AlphaBeta.Examples}{wikipedia-αβ-outcome-and-play}.}

\begin{Verbatim}[commandchars=\\\{\}]
\PYG{n+nf}{optimal\PYGZus{}outcome\PYGZus{}and\PYGZus{}play}\PYG{+w}{ }\PYG{o+ow}{::}\PYG{+w}{ }\PYG{k+kt}{R}
\PYG{n+nf}{optimal\PYGZus{}outcome\PYGZus{}and\PYGZus{}play}\PYG{+w}{ }\PYG{o+ow}{=}
\PYG{+w}{    }\PYG{p}{(}\PYG{n}{runReader}\PYG{+w}{ }\PYG{o}{\PYGZdl{}}\PYG{+w}{ }\PYG{n}{quantifier}\PYG{+w}{ }\PYG{n}{bigPhi}\PYG{+w}{ }\PYG{n}{q}\PYG{p}{)}\PYG{+w}{ }\PYG{p}{(}\PYG{o}{\PYGZhy{}}\PYG{n}{inf}\PYG{p}{,}\PYG{+w}{ }\PYG{n}{inf}\PYG{p}{)}

\PYG{c+c1}{\PYGZhy{}\PYGZhy{} ghci\PYGZgt{} optimal\PYGZus{}outcome\PYGZus{}and\PYGZus{}play}
\PYG{c+c1}{\PYGZhy{}\PYGZhy{} [0,0,0,0] \PYGZhy{}\PYGZgt{} 5}
\PYG{c+c1}{\PYGZhy{}\PYGZhy{} [0,0,0,1] \PYGZhy{}\PYGZgt{} 6}
\PYG{c+c1}{\PYGZhy{}\PYGZhy{} [0,0,0,2] \PYGZhy{}\PYGZgt{} 6}
\PYG{c+c1}{\PYGZhy{}\PYGZhy{} [0,0,1,0] \PYGZhy{}\PYGZgt{} 7}
\PYG{c+c1}{\PYGZhy{}\PYGZhy{} [0,0,1,1] \PYGZhy{}\PYGZgt{} 4}
\PYG{c+c1}{\PYGZhy{}\PYGZhy{} [0,1,0,0] \PYGZhy{}\PYGZgt{} 3}
\PYG{c+c1}{\PYGZhy{}\PYGZhy{} [0,1,0,1] \PYGZhy{}\PYGZgt{} 3}
\PYG{c+c1}{\PYGZhy{}\PYGZhy{} [0,1,0,2] \PYGZhy{}\PYGZgt{} 3}
\PYG{c+c1}{\PYGZhy{}\PYGZhy{} [0,1,1,0] \PYGZhy{}\PYGZgt{} 3}
\PYG{c+c1}{\PYGZhy{}\PYGZhy{} [1,0,0,0] \PYGZhy{}\PYGZgt{} 6}
\PYG{c+c1}{\PYGZhy{}\PYGZhy{} [1,0,0,1] \PYGZhy{}\PYGZgt{} 6}
\PYG{c+c1}{\PYGZhy{}\PYGZhy{} [1,0,0,2] \PYGZhy{}\PYGZgt{} 6}
\PYG{c+c1}{\PYGZhy{}\PYGZhy{} [1,0,1,0] \PYGZhy{}\PYGZgt{} 6}
\PYG{c+c1}{\PYGZhy{}\PYGZhy{} [1,1,0,0] \PYGZhy{}\PYGZgt{} 7}
\PYG{c+c1}{\PYGZhy{}\PYGZhy{} [1,1,0,1] \PYGZhy{}\PYGZgt{} 7}
\PYG{c+c1}{\PYGZhy{}\PYGZhy{} [1,1,0,2] \PYGZhy{}\PYGZgt{} 7}
\PYG{c+c1}{\PYGZhy{}\PYGZhy{} [2,0,0,0] \PYGZhy{}\PYGZgt{} 5}
\PYG{c+c1}{\PYGZhy{}\PYGZhy{} [2,0,1,0] \PYGZhy{}\PYGZgt{} 5}
\PYG{c+c1}{\PYGZhy{}\PYGZhy{} (6,[1,0,0,0])}
\end{Verbatim}
\claim{alpha-beta-optimal-play-example}

\begin{definition} \label{def-optimal-play} A play $t = x_0 x_1 \cdots x_{n-1}$ of a game $G$ is \emph{optimal} if $\out{G_{s_k \cdot x_k}} = \out{G_{s_k}}$ for every $k < n$, where $s_k = x_0 \cdots x_{k-1}$.
\AgdaRefs{\Agda{Games.OptimalPlays}{is-optimal-play}.}
\end{definition}

\begin{corollary} \label{cor-optimal-plays} Let $G'$ be the game $G$ with the outcome function $q'(t) = (q(t), t)$, and with outcomes compared by their first components. Then $\out{G'} = (\out{G}, t)$ for an optimal play $t$ of $G$. Hence, if $\alpha < \out{G} < \beta$, then $\outM{\alpha}{\beta}{G'} = (\out{G}, t)$.
\AgdaRefs{\Agda{Games.AlphaBeta.OptimalPlays}{extended-game-gives-optimal-play},
  \texttt{αβ-search-agrees-with-exhaustive-search};
  \Agda{Games.AlphaBeta.Summary}{Corollary-2}.}
\end{corollary}

\begin{proof} The first claim is by induction on the game tree, since the maximum and the minimum of a non-empty list of outcomes are members of the list, and since the first component of $\out{G'_s}$ is $\out{G_s}$ for every partial play $s$. The second claim follows from the first and Theorem \ref{thm-main}(iii) for $G'$. \qed
\end{proof}

\section{History-Dependent Games}
\label{sec-history-dependent}

So far we have built the alpha-beta search one step at a time. In Section \ref{sec-exhaustive} the exhaustive minimax search of a game tree is the product of the $\min$ and $\max$ quantifiers of its rounds. In Section \ref{sec-alpha-beta} we obtained the alpha-beta search by replacing these quantifiers with monadic ones over the reader monad, without changing how they are composed (using the same product of quantifiers). Section \ref{sec-proof} proved that the two searches compute the same optimal outcome, and Section \ref{sec-optimal-plays} extended both to compute optimal plays as well. In all of these, however, the game has a fixed list of quantifiers. The moves available at a round, and hence the quantifier there, are the same whatever moves were played before. These are \emph{history-free games}.

In this section we go one step further and consider alpha-beta pruning for \emph{history-dependent games}, in which the moves available at a round depend on the moves played so far. Most games of interest are of this kind: in board games, for instance, a move is usually only available if the square or the piece it involves is still free. The monadic quantifiers of Sections \ref{sec-alpha-beta} and \ref{sec-optimal-plays} need no change. What changes is how they are composed: the product of quantifiers becomes the product of history-dependent quantifiers \cite{EO(2010D),EO(2023)}, in which the quantifier at each round is chosen as a function of the partial play leading to it. The correctness results of Sections \ref{sec-proof} and \ref{sec-optimal-plays} carry over to this generality.

We first introduce products of history-dependent quantifiers, and explain why the correctness proofs still apply (Section \ref{sec-history-dependent-products}). We then illustrate the approach with Tic-Tac-Toe, in which each player may only occupy a square that is still free (Section \ref{sec-tic-tac-toe}). We use it to compute the optimal outcome and an optimal play, and compare the number of plays evaluated by the alpha-beta search with that of the exhaustive search.

\subsection{Products of history-dependent quantifiers}
\label{sec-history-dependent-products}

In the games considered so far, the quantifier at each round is fixed in advance. In many games, such as Tic-Tac-Toe, the moves available at a round depend on the moves played so far. Following \cite{EO(2010D),EO(2023)}, we model such a game by letting the quantifier at each round depend on the partial play that leads to it, so that a game is given by quantifiers
\[
\phi_i \colon X^i \to K_R X \qquad (i < n)
\]
together with an outcome function $q \colon X^n \to R$. The product of quantifiers is then defined for every partial play $s$ of length $i \leq n$ by
\[
(\bigotimes\nolimits_s \phi)(q) \fdef
\begin{cases}
q(s) & \text{if } i = n, \\
\phi_i(s)(\lambda x . (\bigotimes_{s \cdot x} \phi)(q)) & \text{if } i < n,
\end{cases}
\]
and the optimal outcome is $\out{G} \fdef (\bigotimes_{[\,]} \phi)(q)$, as computed by the function \Verb[commandchars=\\\{\}]|\PYG{n+nf}{hsequence}| of Appendix \ref{app-k}. The monadic quantifiers $\phi_i^* \colon X^i \to K_{TR} X$ give the monadic outcome $\outM{\alpha}{\beta}{G}$ in the same way.

For the results of Sections \ref{sec-proof} and \ref{sec-optimal-plays}, each $\phi_i(s)$ is $\phi_{\max}$ or $\phi_{\min}$ applied to a non-empty list of the moves available at $s$, and $\phi_i^*(s)$ is the corresponding $\phi_{\max}^*$ or $\phi_{\min}^*$. With $\phi$ the quantifier $\phi_i(s)$ at a partial play $s$ of length $i$, the definitions of $\out{G_s}$ and $\outM{\alpha}{\beta}{G_s}$, from Section \ref{sec-proof}, should then use the dependent product of quantifiers. The proofs of Theorem \ref{thm-main} and Corollaries \ref{cor-bounds} and \ref{cor-optimal-plays} are by induction on the game tree, and at each position they use the properties of the quantifier there, and so they apply to history-dependent games without change.

In \cite{EO(2023)}, to which we refer the reader for the details, the type of moves at a round, and the number of rounds, may also depend on the moves played so far. A game is then given by a tree of types, whose paths are the plays, together with a quantifier at each internal node and an outcome function on the paths. Our Agda formalization is developed in this generality, with each internal node labelled by a player and the non-empty list of the moves available there, and Theorem \ref{thm-main} and Corollaries \ref{cor-bounds} and \ref{cor-optimal-plays} are proved for all such games.
\AgdaRefs{\Agda{Games.TypeTrees}{𝑻}, \texttt{Path}; \Agda{Games.AlphaBeta.Player}{Node};
  \Agda{Games.AlphaBeta.ViaProductsOfQuantifiers}{quantifier-tree}, \texttt{theorem}.}

\subsection{Alpha-beta search of Tic-Tac-Toe}
\label{sec-tic-tac-toe}

In Tic-Tac-Toe, the moves are the nine squares of the board, numbered $0, \ldots, 8$ row by row, and a partial play $s$ is the list of the squares occupied so far, in the order in which they were occupied. The first player maximises and moves at the even rounds, the second minimises and moves at the odd rounds, and each may choose any square not yet occupied. Thus $\phi_i(s)$ is $\phi_{\max}(xs)$ for even $i$ and $\phi_{\min}(xs)$ for odd $i$, where $xs$ is the list of the squares not in $s$. We use the $\min$ and $\max$ quantifiers of Section \ref{sec-optimal-plays} to calculate (with alpha-beta pruning) the optimal outcome and an optimal play:

\begin{Verbatim}[commandchars=\\\{\}]
\PYG{k+kr}{type}\PYG{+w}{ }\PYG{k+kt}{Move}\PYG{+w}{ }\PYG{o+ow}{=}\PYG{+w}{ }\PYG{k+kt}{Int}
\PYG{k+kr}{type}\PYG{+w}{ }\PYG{k+kt}{Path}\PYG{+w}{ }\PYG{o+ow}{=}\PYG{+w}{ }\PYG{p}{[}\PYG{k+kt}{Move}\PYG{p}{]}
\PYG{k+kr}{type}\PYG{+w}{ }\PYG{k+kt}{Moves}\PYG{+w}{ }\PYG{o+ow}{=}\PYG{+w}{ }\PYG{p}{[}\PYG{k+kt}{Int}\PYG{p}{]}

\PYG{k+kr}{type}\PYG{+w}{ }\PYG{k+kt}{Board}\PYG{+w}{ }\PYG{o+ow}{=}\PYG{+w}{ }\PYG{p}{([}\PYG{k+kt}{Move}\PYG{p}{],}\PYG{+w}{ }\PYG{p}{[}\PYG{k+kt}{Move}\PYG{p}{])}
\PYG{k+kr}{data}\PYG{+w}{ }\PYG{k+kt}{Player}\PYG{+w}{ }\PYG{o+ow}{=}\PYG{+w}{ }\PYG{k+kt}{X}\PYG{+w}{ }\PYG{o}{|}\PYG{+w}{ }\PYG{k+kt}{O}

\PYG{k+kr}{type}\PYG{+w}{ }\PYG{k+kt}{Three}\PYG{+w}{ }\PYG{o+ow}{=}\PYG{+w}{ }\PYG{k+kt}{Int}

\PYG{k+kr}{type}\PYG{+w}{ }\PYG{k+kt}{R}\PYG{+w}{ }\PYG{o+ow}{=}\PYG{+w}{ }\PYG{p}{(}\PYG{k+kt}{Three}\PYG{p}{,}\PYG{+w}{ }\PYG{k+kt}{Path}\PYG{p}{)}

\PYG{k+kr}{type}\PYG{+w}{ }\PYG{k+kt}{T}\PYG{+w}{ }\PYG{o+ow}{=}\PYG{+w}{ }\PYG{k+kt}{Reader}\PYG{+w}{ }\PYG{p}{(}\PYG{k+kt}{Int}\PYG{p}{,}\PYG{+w}{ }\PYG{k+kt}{Int}\PYG{p}{)}\PYG{+w}{ }\PYG{c+c1}{\PYGZhy{}\PYGZhy{} Reader Monad}

\PYG{c+c1}{\PYGZhy{}\PYGZhy{} In this case +1 and \PYGZhy{}1 can work as upper and lower}
\PYG{c+c1}{\PYGZhy{}\PYGZhy{} bounds for the possible game outcomes}
\PYG{n+nf}{inf}\PYG{+w}{ }\PYG{o+ow}{::}\PYG{+w}{ }\PYG{k+kt}{Int}
\PYG{n+nf}{inf}\PYG{+w}{ }\PYG{o+ow}{=}\PYG{+w}{ }\PYG{l+m+mi}{1}

\PYG{c+c1}{\PYGZhy{}\PYGZhy{} Defining the outcome function q for Tic\PYGZhy{}Tac\PYGZhy{}Toe}

\PYG{n+nf}{wins}\PYG{+w}{ }\PYG{o+ow}{::}\PYG{+w}{ }\PYG{k+kt}{Path}\PYG{+w}{ }\PYG{o+ow}{\PYGZhy{}\PYGZgt{}}\PYG{+w}{ }\PYG{k+kt}{Bool}
\PYG{n+nf}{wins}\PYG{+w}{ }\PYG{n}{xs}\PYG{+w}{ }\PYG{o+ow}{=}\PYG{+w}{ }\PYG{n}{some}
\PYG{+w}{  }\PYG{p}{[[}\PYG{l+m+mi}{0}\PYG{p}{,}\PYG{l+m+mi}{1}\PYG{p}{,}\PYG{l+m+mi}{2}\PYG{p}{],[}\PYG{l+m+mi}{3}\PYG{p}{,}\PYG{l+m+mi}{4}\PYG{p}{,}\PYG{l+m+mi}{5}\PYG{p}{],[}\PYG{l+m+mi}{6}\PYG{p}{,}\PYG{l+m+mi}{7}\PYG{p}{,}\PYG{l+m+mi}{8}\PYG{p}{],[}\PYG{l+m+mi}{0}\PYG{p}{,}\PYG{l+m+mi}{3}\PYG{p}{,}\PYG{l+m+mi}{6}\PYG{p}{],[}\PYG{l+m+mi}{1}\PYG{p}{,}\PYG{l+m+mi}{4}\PYG{p}{,}\PYG{l+m+mi}{7}\PYG{p}{],[}\PYG{l+m+mi}{2}\PYG{p}{,}\PYG{l+m+mi}{5}\PYG{p}{,}\PYG{l+m+mi}{8}\PYG{p}{],[}\PYG{l+m+mi}{0}\PYG{p}{,}\PYG{l+m+mi}{4}\PYG{p}{,}\PYG{l+m+mi}{8}\PYG{p}{],[}\PYG{l+m+mi}{2}\PYG{p}{,}\PYG{l+m+mi}{4}\PYG{p}{,}\PYG{l+m+mi}{6}\PYG{p}{]]}
\PYG{+w}{    }\PYG{p}{(}\PYG{n+nf}{\PYGZbs{}}\PYG{n}{ys}\PYG{+w}{ }\PYG{o+ow}{\PYGZhy{}\PYGZgt{}}\PYG{+w}{ }\PYG{n}{contained}\PYG{+w}{ }\PYG{n}{ys}\PYG{+w}{ }\PYG{n}{xs}\PYG{p}{)}

\PYG{n+nf}{value}\PYG{+w}{ }\PYG{o+ow}{::}\PYG{+w}{ }\PYG{k+kt}{Board}\PYG{+w}{ }\PYG{o+ow}{\PYGZhy{}\PYGZgt{}}\PYG{+w}{ }\PYG{k+kt}{Three}
\PYG{n+nf}{value}\PYG{+w}{ }\PYG{p}{(}\PYG{n}{x}\PYG{p}{,}\PYG{n}{o}\PYG{p}{)}\PYG{+w}{ }\PYG{o+ow}{=}\PYG{+w}{ }\PYG{k+kr}{if}\PYG{+w}{ }\PYG{n}{wins}\PYG{+w}{ }\PYG{n}{x}\PYG{+w}{ }\PYG{k+kr}{then}\PYG{+w}{ }\PYG{l+m+mi}{1}\PYG{+w}{ }\PYG{k+kr}{else}\PYG{+w}{ }\PYG{k+kr}{if}\PYG{+w}{ }\PYG{n}{wins}\PYG{+w}{ }\PYG{n}{o}\PYG{+w}{ }\PYG{k+kr}{then}\PYG{+w}{ }\PYG{o}{\PYGZhy{}}\PYG{l+m+mi}{1}\PYG{+w}{ }\PYG{k+kr}{else}\PYG{+w}{ }\PYG{l+m+mi}{0}

\PYG{n+nf}{outcome}\PYG{+w}{ }\PYG{o+ow}{::}\PYG{+w}{ }\PYG{k+kt}{Player}\PYG{+w}{ }\PYG{o+ow}{\PYGZhy{}\PYGZgt{}}\PYG{+w}{ }\PYG{p}{[}\PYG{k+kt}{Move}\PYG{p}{]}\PYG{+w}{ }\PYG{o+ow}{\PYGZhy{}\PYGZgt{}}\PYG{+w}{ }\PYG{k+kt}{Board}\PYG{+w}{ }\PYG{o+ow}{\PYGZhy{}\PYGZgt{}}\PYG{+w}{ }\PYG{k+kt}{Board}
\PYG{n+nf}{outcome}\PYG{+w}{ }\PYG{n}{whoever}\PYG{+w}{ }\PYG{k+kt}{[]}\PYG{+w}{ }\PYG{n}{board}\PYG{+w}{ }\PYG{o+ow}{=}\PYG{+w}{ }\PYG{n}{board}
\PYG{n+nf}{outcome}\PYG{+w}{ }\PYG{k+kt}{X}\PYG{+w}{ }\PYG{p}{(}\PYG{n}{m}\PYG{+w}{ }\PYG{k+kt}{:}\PYG{+w}{ }\PYG{n}{ms}\PYG{p}{)}\PYG{+w}{ }\PYG{p}{(}\PYG{n}{x}\PYG{p}{,}\PYG{+w}{ }\PYG{n}{o}\PYG{p}{)}\PYG{+w}{ }\PYG{o+ow}{=}
\PYG{+w}{  }\PYG{k+kr}{if}\PYG{+w}{ }\PYG{n}{wins}\PYG{+w}{ }\PYG{n}{o}\PYG{+w}{ }\PYG{k+kr}{then}\PYG{+w}{ }\PYG{p}{(}\PYG{n}{x}\PYG{p}{,}\PYG{+w}{ }\PYG{n}{o}\PYG{p}{)}\PYG{+w}{ }\PYG{k+kr}{else}\PYG{+w}{ }\PYG{n}{outcome}\PYG{+w}{ }\PYG{k+kt}{O}\PYG{+w}{ }\PYG{n}{ms}\PYG{+w}{ }\PYG{p}{(}\PYG{n}{m}\PYG{+w}{ }\PYG{k+kt}{:}\PYG{+w}{ }\PYG{n}{x}\PYG{p}{,}\PYG{+w}{ }\PYG{n}{o}\PYG{p}{)}
\PYG{n+nf}{outcome}\PYG{+w}{ }\PYG{k+kt}{O}\PYG{+w}{ }\PYG{p}{(}\PYG{n}{m}\PYG{+w}{ }\PYG{k+kt}{:}\PYG{+w}{ }\PYG{n}{ms}\PYG{p}{)}\PYG{+w}{ }\PYG{p}{(}\PYG{n}{x}\PYG{p}{,}\PYG{+w}{ }\PYG{n}{o}\PYG{p}{)}\PYG{+w}{ }\PYG{o+ow}{=}
\PYG{+w}{  }\PYG{k+kr}{if}\PYG{+w}{ }\PYG{n}{wins}\PYG{+w}{ }\PYG{n}{x}\PYG{+w}{ }\PYG{k+kr}{then}\PYG{+w}{ }\PYG{p}{(}\PYG{n}{x}\PYG{p}{,}\PYG{+w}{ }\PYG{n}{o}\PYG{p}{)}\PYG{+w}{ }\PYG{k+kr}{else}\PYG{+w}{ }\PYG{n}{outcome}\PYG{+w}{ }\PYG{k+kt}{X}\PYG{+w}{ }\PYG{n}{ms}\PYG{+w}{ }\PYG{p}{(}\PYG{n}{x}\PYG{p}{,}\PYG{+w}{ }\PYG{n}{m}\PYG{+w}{ }\PYG{k+kt}{:}\PYG{+w}{ }\PYG{n}{o}\PYG{p}{)}

\PYG{n+nf}{q\PYGZsq{}}\PYG{+w}{ }\PYG{o+ow}{::}\PYG{+w}{ }\PYG{k+kt}{Path}\PYG{+w}{ }\PYG{o+ow}{\PYGZhy{}\PYGZgt{}}\PYG{+w}{ }\PYG{k+kt}{Three}
\PYG{n+nf}{q\PYGZsq{}}\PYG{+w}{ }\PYG{n}{ms}\PYG{+w}{ }\PYG{o+ow}{=}\PYG{+w}{ }\PYG{n}{value}\PYG{+w}{ }\PYG{o}{\PYGZdl{}}\PYG{+w}{ }\PYG{n}{outcome}\PYG{+w}{ }\PYG{k+kt}{X}\PYG{+w}{ }\PYG{n}{ms}\PYG{+w}{ }\PYG{p}{(}\PYG{k+kt}{[]}\PYG{p}{,}\PYG{k+kt}{[]}\PYG{p}{)}

\PYG{n+nf}{q}\PYG{+w}{ }\PYG{o+ow}{::}\PYG{+w}{ }\PYG{k+kt}{Path}\PYG{+w}{ }\PYG{o+ow}{\PYGZhy{}\PYGZgt{}}\PYG{+w}{ }\PYG{k+kt}{T}\PYG{+w}{ }\PYG{k+kt}{R}
\PYG{n+nf}{q}\PYG{+w}{ }\PYG{n}{xs}\PYG{+w}{ }\PYG{o+ow}{=}\PYG{+w}{ }\PYG{n}{return}\PYG{+w}{ }\PYG{o}{\PYGZdl{}}\PYG{+w}{ }\PYG{k+kr}{let}\PYG{+w}{ }\PYG{n}{r}\PYG{+w}{ }\PYG{o+ow}{=}\PYG{+w}{ }\PYG{n}{q\PYGZsq{}}\PYG{+w}{ }\PYG{n}{xs}\PYG{+w}{ }\PYG{k+kr}{in}
\PYG{+w}{  }\PYG{n}{trace}\PYG{+w}{ }\PYG{p}{(}\PYG{n}{show}\PYG{+w}{ }\PYG{n}{xs}\PYG{+w}{ }\PYG{o}{++}\PYG{+w}{ }\PYG{l+s}{\PYGZdq{} \PYGZhy{}\PYGZgt{} \PYGZdq{}}\PYG{+w}{ }\PYG{o}{++}\PYG{+w}{ }\PYG{n}{show}\PYG{+w}{ }\PYG{n}{r}\PYG{p}{)}\PYG{+w}{ }\PYG{p}{(}\PYG{n}{r}\PYG{p}{,}\PYG{+w}{ }\PYG{n}{xs}\PYG{p}{)}

\PYG{c+c1}{\PYGZhy{}\PYGZhy{} Defining the (history\PYGZhy{}dependent) quantifiers}

\PYG{c+c1}{\PYGZhy{}\PYGZhy{} game quantifiers for max\PYGZhy{}min players}
\PYG{n+nf}{phisT}\PYG{+w}{ }\PYG{o+ow}{::}\PYG{+w}{ }\PYG{p}{[}\PYG{k+kt}{Path}\PYG{+w}{ }\PYG{o+ow}{\PYGZhy{}\PYGZgt{}}\PYG{+w}{ }\PYG{k+kt}{K}\PYG{+w}{ }\PYG{p}{(}\PYG{k+kt}{T}\PYG{+w}{ }\PYG{k+kt}{R}\PYG{p}{)}\PYG{+w}{ }\PYG{k+kt}{Move}\PYG{p}{]}
\PYG{n+nf}{phisT}\PYG{+w}{ }\PYG{o+ow}{=}\PYG{+w}{ }\PYG{p}{(}\PYG{n}{phiMax}\PYG{+w}{ }\PYG{o}{.}\PYG{+w}{ }\PYG{n}{minus}\PYG{+w}{ }\PYG{p}{[}\PYG{l+m+mi}{0}\PYG{o}{..}\PYG{l+m+mi}{8}\PYG{p}{])}\PYG{+w}{ }\PYG{k+kt}{:}\PYG{+w}{ }\PYG{p}{(}\PYG{n}{phiMin}\PYG{+w}{ }\PYG{o}{.}\PYG{+w}{ }\PYG{n}{minus}\PYG{+w}{ }\PYG{p}{[}\PYG{l+m+mi}{0}\PYG{o}{..}\PYG{l+m+mi}{8}\PYG{p}{])}\PYG{+w}{ }\PYG{k+kt}{:}\PYG{+w}{ }\PYG{n}{phisT}

\PYG{c+c1}{\PYGZhy{}\PYGZhy{} global quantifier for the game tree}
\PYG{c+c1}{\PYGZhy{}\PYGZhy{} we use hsequence since the available moves at any point in}
\PYG{c+c1}{\PYGZhy{}\PYGZhy{} the game depend on what has been played before}
\PYG{n+nf}{bigPhi}\PYG{+w}{ }\PYG{o+ow}{::}\PYG{+w}{ }\PYG{k+kt}{K}\PYG{+w}{ }\PYG{p}{(}\PYG{k+kt}{T}\PYG{+w}{ }\PYG{k+kt}{R}\PYG{p}{)}\PYG{+w}{ }\PYG{k+kt}{Path}
\PYG{n+nf}{bigPhi}\PYG{+w}{ }\PYG{o+ow}{=}\PYG{+w}{ }\PYG{n}{hsequence}\PYG{+w}{ }\PYG{o}{\PYGZdl{}}\PYG{+w}{ }\PYG{n}{take}\PYG{+w}{ }\PYG{l+m+mi}{9}\PYG{+w}{ }\PYG{n}{phisT}

\PYG{c+c1}{\PYGZhy{}\PYGZhy{} Calculating optimal outcome and an optimal play}
\PYG{n+nf}{optimal\PYGZus{}outcome\PYGZus{}and\PYGZus{}play}\PYG{+w}{ }\PYG{o+ow}{::}\PYG{+w}{ }\PYG{k+kt}{R}
\PYG{n+nf}{optimal\PYGZus{}outcome\PYGZus{}and\PYGZus{}play}\PYG{+w}{ }\PYG{o+ow}{=}\PYG{+w}{ }\PYG{p}{(}\PYG{n}{runReader}\PYG{+w}{ }\PYG{o}{\PYGZdl{}}\PYG{+w}{ }\PYG{n}{quantifier}\PYG{+w}{ }\PYG{n}{bigPhi}\PYG{+w}{ }\PYG{n}{q}\PYG{p}{)}\PYG{+w}{ }\PYG{p}{(}\PYG{o}{\PYGZhy{}}\PYG{n}{inf}\PYG{p}{,}\PYG{+w}{ }\PYG{n}{inf}\PYG{p}{)}

\PYG{c+c1}{\PYGZhy{}\PYGZhy{} ghci\PYGZgt{} optimal\PYGZus{}outcome\PYGZus{}and\PYGZus{}play}
\PYG{c+c1}{\PYGZhy{}\PYGZhy{} [0,1,2,3,4,5,6,7,8] \PYGZhy{}\PYGZgt{} 1}
\PYG{c+c1}{\PYGZhy{}\PYGZhy{} [0,1,2,3,4,5,6,8,7] \PYGZhy{}\PYGZgt{} 1}
\PYG{c+c1}{\PYGZhy{}\PYGZhy{} [0,1,2,3,4,6,5,7,8] \PYGZhy{}\PYGZgt{} 1}
\PYG{c+c1}{\PYGZhy{}\PYGZhy{} ...}
\PYG{c+c1}{\PYGZhy{}\PYGZhy{} [8,4,7,6,5,2,1,0,3] \PYGZhy{}\PYGZgt{} \PYGZhy{}1}
\PYG{c+c1}{\PYGZhy{}\PYGZhy{} [8,4,7,6,5,2,3,0,1] \PYGZhy{}\PYGZgt{} \PYGZhy{}1}
\PYG{c+c1}{\PYGZhy{}\PYGZhy{} (0,[0,4,1,2,6,3,5,7,8])}
\end{Verbatim}

Note that we are taking the initial values of $\alpha$ and $\beta$ to be $-1$ and $1$, respectively. Since the bounds $-1$ and $1$ are also outcomes of the game, Corollary \ref{cor-bounds}(a) does not apply. But the computed outcome $0$ lies strictly between them, so it is the optimal outcome, by Corollary \ref{cor-bounds}(b).

In this Tic-Tac-Toe example the exhaustive search evaluates the outcome function on all $9! = 362\,880$ full plays of the game tree, whereas the alpha-beta search evaluates it on only $9\,335$ of them, about 39 times fewer, and computes the same optimal outcome and optimal play:
\begin{verbatim}
(0,[0,4,1,2,6,3,5,7,8])
\end{verbatim}

\phantomsection

\appendix
\renewcommand{\theHsection}{appendix.\Alph{section}}

\section{Haskell -- Utils}
\label{app-utils}

\begin{Verbatim}[commandchars=\\\{\}]
\PYG{k+kr}{module}\PYG{+w}{ }\PYG{n+nn}{Utils}\PYG{+w}{ }\PYG{k+kr}{where}

\PYG{n+nf}{takeUntil}\PYG{+w}{ }\PYG{o+ow}{::}\PYG{+w}{ }\PYG{p}{(}\PYG{n}{a}\PYG{+w}{ }\PYG{o+ow}{\PYGZhy{}\PYGZgt{}}\PYG{+w}{ }\PYG{k+kt}{Bool}\PYG{p}{)}\PYG{+w}{ }\PYG{o+ow}{\PYGZhy{}\PYGZgt{}}\PYG{+w}{ }\PYG{p}{[}\PYG{n}{a}\PYG{p}{]}\PYG{+w}{ }\PYG{o+ow}{\PYGZhy{}\PYGZgt{}}\PYG{+w}{ }\PYG{p}{[}\PYG{n}{a}\PYG{p}{]}
\PYG{n+nf}{takeUntil}\PYG{+w}{ }\PYG{k+kr}{\PYGZus{}}\PYG{+w}{ }\PYG{k+kt}{[]}\PYG{+w}{ }\PYG{o+ow}{=}\PYG{+w}{ }\PYG{k+kt}{[]}
\PYG{n+nf}{takeUntil}\PYG{+w}{ }\PYG{n}{p}\PYG{+w}{ }\PYG{p}{(}\PYG{n}{x}\PYG{k+kt}{:}\PYG{n}{xs}\PYG{p}{)}\PYG{+w}{ }\PYG{o+ow}{=}\PYG{+w}{ }\PYG{n}{x}\PYG{+w}{ }\PYG{k+kt}{:}\PYG{+w}{ }\PYG{k+kr}{if}\PYG{+w}{ }\PYG{n}{p}\PYG{+w}{ }\PYG{n}{x}\PYG{+w}{ }\PYG{k+kr}{then}\PYG{+w}{ }\PYG{k+kt}{[]}\PYG{+w}{ }\PYG{k+kr}{else}\PYG{+w}{ }\PYG{n}{takeUntil}\PYG{+w}{ }\PYG{n}{p}\PYG{+w}{ }\PYG{n}{xs}

\PYG{n+nf}{pickMax}\PYG{+w}{ }\PYG{o+ow}{::}\PYG{+w}{ }\PYG{p}{(}\PYG{k+kt}{Ord}\PYG{+w}{ }\PYG{n}{a}\PYG{p}{)}\PYG{+w}{ }\PYG{o+ow}{=\PYGZgt{}}\PYG{+w}{ }\PYG{p}{[(}\PYG{n}{a}\PYG{p}{,}\PYG{+w}{ }\PYG{n}{b}\PYG{p}{)]}\PYG{+w}{ }\PYG{o+ow}{\PYGZhy{}\PYGZgt{}}\PYG{+w}{ }\PYG{p}{(}\PYG{n}{a}\PYG{p}{,}\PYG{+w}{ }\PYG{n}{b}\PYG{p}{)}
\PYG{n+nf}{pickMax}\PYG{+w}{ }\PYG{p}{[}\PYG{n}{v}\PYG{p}{]}\PYG{+w}{ }\PYG{o+ow}{=}\PYG{+w}{ }\PYG{n}{v}
\PYG{n+nf}{pickMax}\PYG{+w}{ }\PYG{p}{((}\PYG{n}{v}\PYG{p}{,}\PYG{+w}{ }\PYG{n}{x}\PYG{p}{)}\PYG{+w}{ }\PYG{k+kt}{:}\PYG{+w}{ }\PYG{p}{(}\PYG{n}{w}\PYG{p}{,}\PYG{+w}{ }\PYG{n}{y}\PYG{p}{)}\PYG{+w}{ }\PYG{k+kt}{:}\PYG{+w}{ }\PYG{n}{zs}\PYG{p}{)}\PYG{+w}{ }\PYG{o+ow}{=}
\PYG{+w}{  }\PYG{k+kr}{if}\PYG{+w}{ }\PYG{n}{v}\PYG{+w}{ }\PYG{o}{\PYGZgt{}=}\PYG{+w}{ }\PYG{n}{w}\PYG{+w}{ }\PYG{k+kr}{then}\PYG{+w}{ }\PYG{n}{pickMax}\PYG{+w}{ }\PYG{p}{((}\PYG{n}{v}\PYG{p}{,}\PYG{+w}{ }\PYG{n}{x}\PYG{p}{)}\PYG{+w}{ }\PYG{k+kt}{:}\PYG{+w}{ }\PYG{n}{zs}\PYG{p}{)}\PYG{+w}{ }\PYG{k+kr}{else}\PYG{+w}{ }\PYG{n}{pickMax}\PYG{+w}{ }\PYG{p}{((}\PYG{n}{w}\PYG{p}{,}\PYG{+w}{ }\PYG{n}{y}\PYG{p}{)}\PYG{+w}{ }\PYG{k+kt}{:}\PYG{+w}{ }\PYG{n}{zs}\PYG{p}{)}

\PYG{n+nf}{pickMin}\PYG{+w}{ }\PYG{o+ow}{::}\PYG{+w}{ }\PYG{p}{(}\PYG{k+kt}{Ord}\PYG{+w}{ }\PYG{n}{a}\PYG{p}{)}\PYG{+w}{ }\PYG{o+ow}{=\PYGZgt{}}\PYG{+w}{ }\PYG{p}{[(}\PYG{n}{a}\PYG{p}{,}\PYG{+w}{ }\PYG{n}{b}\PYG{p}{)]}\PYG{+w}{ }\PYG{o+ow}{\PYGZhy{}\PYGZgt{}}\PYG{+w}{ }\PYG{p}{(}\PYG{n}{a}\PYG{p}{,}\PYG{+w}{ }\PYG{n}{b}\PYG{p}{)}
\PYG{n+nf}{pickMin}\PYG{+w}{ }\PYG{p}{[}\PYG{n}{v}\PYG{p}{]}\PYG{+w}{ }\PYG{o+ow}{=}\PYG{+w}{ }\PYG{n}{v}
\PYG{n+nf}{pickMin}\PYG{+w}{ }\PYG{p}{((}\PYG{n}{v}\PYG{p}{,}\PYG{+w}{ }\PYG{n}{x}\PYG{p}{)}\PYG{+w}{ }\PYG{k+kt}{:}\PYG{+w}{ }\PYG{p}{(}\PYG{n}{w}\PYG{p}{,}\PYG{+w}{ }\PYG{n}{y}\PYG{p}{)}\PYG{+w}{ }\PYG{k+kt}{:}\PYG{+w}{ }\PYG{n}{zs}\PYG{p}{)}\PYG{+w}{ }\PYG{o+ow}{=}
\PYG{+w}{  }\PYG{k+kr}{if}\PYG{+w}{ }\PYG{n}{v}\PYG{+w}{ }\PYG{o}{\PYGZlt{}=}\PYG{+w}{ }\PYG{n}{w}\PYG{+w}{ }\PYG{k+kr}{then}\PYG{+w}{ }\PYG{n}{pickMin}\PYG{+w}{ }\PYG{p}{((}\PYG{n}{v}\PYG{p}{,}\PYG{+w}{ }\PYG{n}{x}\PYG{p}{)}\PYG{+w}{ }\PYG{k+kt}{:}\PYG{+w}{ }\PYG{n}{zs}\PYG{p}{)}\PYG{+w}{ }\PYG{k+kr}{else}\PYG{+w}{ }\PYG{n}{pickMin}\PYG{+w}{ }\PYG{p}{((}\PYG{n}{w}\PYG{p}{,}\PYG{+w}{ }\PYG{n}{y}\PYG{p}{)}\PYG{+w}{ }\PYG{k+kt}{:}\PYG{+w}{ }\PYG{n}{zs}\PYG{p}{)}

\PYG{n+nf}{some}\PYG{p}{,}\PYG{+w}{ }\PYG{n}{every}\PYG{+w}{  }\PYG{o+ow}{::}\PYG{+w}{ }\PYG{p}{[}\PYG{n}{a}\PYG{p}{]}\PYG{+w}{ }\PYG{o+ow}{\PYGZhy{}\PYGZgt{}}\PYG{+w}{ }\PYG{p}{(}\PYG{n}{a}\PYG{+w}{ }\PYG{o+ow}{\PYGZhy{}\PYGZgt{}}\PYG{+w}{ }\PYG{k+kt}{Bool}\PYG{p}{)}\PYG{+w}{ }\PYG{o+ow}{\PYGZhy{}\PYGZgt{}}\PYG{+w}{ }\PYG{k+kt}{Bool}
\PYG{n+nf}{some}\PYG{+w}{  }\PYG{n}{xs}\PYG{+w}{ }\PYG{n}{p}\PYG{+w}{ }\PYG{o+ow}{=}\PYG{+w}{ }\PYG{n}{any}\PYG{+w}{ }\PYG{n}{p}\PYG{+w}{ }\PYG{n}{xs}
\PYG{n+nf}{every}\PYG{+w}{ }\PYG{n}{xs}\PYG{+w}{ }\PYG{n}{p}\PYG{+w}{ }\PYG{o+ow}{=}\PYG{+w}{ }\PYG{n}{all}\PYG{+w}{ }\PYG{n}{p}\PYG{+w}{ }\PYG{n}{xs}

\PYG{n+nf}{contained}\PYG{+w}{ }\PYG{o+ow}{::}\PYG{+w}{ }\PYG{k+kt}{Eq}\PYG{+w}{ }\PYG{n}{x}\PYG{+w}{ }\PYG{o+ow}{=\PYGZgt{}}\PYG{+w}{ }\PYG{p}{[}\PYG{n}{x}\PYG{p}{]}\PYG{+w}{ }\PYG{o+ow}{\PYGZhy{}\PYGZgt{}}\PYG{+w}{ }\PYG{p}{[}\PYG{n}{x}\PYG{p}{]}\PYG{+w}{ }\PYG{o+ow}{\PYGZhy{}\PYGZgt{}}\PYG{+w}{ }\PYG{k+kt}{Bool}
\PYG{n+nf}{contained}\PYG{+w}{ }\PYG{n}{xs}\PYG{+w}{ }\PYG{n}{ys}\PYG{+w}{ }\PYG{o+ow}{=}\PYG{+w}{ }\PYG{n}{every}\PYG{+w}{ }\PYG{n}{xs}\PYG{+w}{ }\PYG{p}{(}\PYG{n+nf}{\PYGZbs{}}\PYG{n}{x}\PYG{+w}{ }\PYG{o+ow}{\PYGZhy{}\PYGZgt{}}\PYG{+w}{ }\PYG{n}{elem}\PYG{+w}{ }\PYG{n}{x}\PYG{+w}{ }\PYG{n}{ys}\PYG{p}{)}

\PYG{n+nf}{minus}\PYG{+w}{ }\PYG{o+ow}{::}\PYG{+w}{ }\PYG{k+kt}{Eq}\PYG{+w}{ }\PYG{n}{x}\PYG{+w}{ }\PYG{o+ow}{=\PYGZgt{}}\PYG{+w}{ }\PYG{p}{[}\PYG{n}{x}\PYG{p}{]}\PYG{+w}{ }\PYG{o+ow}{\PYGZhy{}\PYGZgt{}}\PYG{+w}{ }\PYG{p}{[}\PYG{n}{x}\PYG{p}{]}\PYG{+w}{ }\PYG{o+ow}{\PYGZhy{}\PYGZgt{}}\PYG{+w}{ }\PYG{p}{[}\PYG{n}{x}\PYG{p}{]}
\PYG{n+nf}{minus}\PYG{+w}{ }\PYG{n}{xs}\PYG{+w}{ }\PYG{n}{ys}\PYG{+w}{ }\PYG{o+ow}{=}\PYG{+w}{ }\PYG{p}{[}\PYG{n}{x}\PYG{+w}{ }\PYG{o}{|}\PYG{+w}{ }\PYG{n}{x}\PYG{o+ow}{\PYGZlt{}\PYGZhy{}}\PYG{n}{xs}\PYG{p}{,}\PYG{+w}{ }\PYG{n}{not}\PYG{p}{(}\PYG{n}{elem}\PYG{+w}{ }\PYG{n}{x}\PYG{+w}{ }\PYG{n}{ys}\PYG{p}{)]}
\end{Verbatim}

\section{Haskell -- Continuation Monad}
\label{app-k}

\begin{Verbatim}[commandchars=\\\{\}]
\PYG{k+kr}{module}\PYG{+w}{ }\PYG{n+nn}{K}\PYG{+w}{ }\PYG{k+kr}{where}

\PYG{k+kr}{data}\PYG{+w}{ }\PYG{k+kt}{K}\PYG{+w}{ }\PYG{n}{r}\PYG{+w}{ }\PYG{n}{x}\PYG{+w}{ }\PYG{o+ow}{=}\PYG{+w}{ }\PYG{k+kt}{K}\PYG{+w}{ }\PYG{p}{\PYGZob{}}\PYG{n}{quantifier}\PYG{+w}{ }\PYG{o+ow}{::}\PYG{+w}{ }\PYG{p}{(}\PYG{n}{x}\PYG{+w}{ }\PYG{o+ow}{\PYGZhy{}\PYGZgt{}}\PYG{+w}{ }\PYG{n}{r}\PYG{p}{)}\PYG{+w}{ }\PYG{o+ow}{\PYGZhy{}\PYGZgt{}}\PYG{+w}{ }\PYG{n}{r}\PYG{p}{\PYGZcb{}}

\PYG{k+kr}{instance}\PYG{+w}{ }\PYG{k+kt}{Functor}\PYG{+w}{ }\PYG{p}{(}\PYG{k+kt}{K}\PYG{+w}{ }\PYG{n}{r}\PYG{p}{)}\PYG{+w}{ }\PYG{k+kr}{where}
\PYG{+w}{    }\PYG{n}{fmap}\PYG{+w}{ }\PYG{n}{f}\PYG{+w}{ }\PYG{n}{phi}\PYG{+w}{ }\PYG{o+ow}{=}\PYG{+w}{ }\PYG{k+kt}{K}\PYG{+w}{ }\PYG{p}{(}\PYG{n+nf}{\PYGZbs{}}\PYG{n}{p}\PYG{+w}{ }\PYG{o+ow}{\PYGZhy{}\PYGZgt{}}\PYG{+w}{ }\PYG{n}{quantifier}\PYG{+w}{ }\PYG{n}{phi}\PYG{+w}{ }\PYG{o}{\PYGZdl{}}\PYG{+w}{ }\PYG{n}{p}\PYG{+w}{ }\PYG{o}{.}\PYG{+w}{ }\PYG{n}{f}\PYG{p}{)}

\PYG{k+kr}{instance}\PYG{+w}{ }\PYG{k+kt}{Applicative}\PYG{+w}{ }\PYG{p}{(}\PYG{k+kt}{K}\PYG{+w}{ }\PYG{n}{r}\PYG{p}{)}\PYG{+w}{ }\PYG{k+kr}{where}
\PYG{+w}{    }\PYG{n}{pure}\PYG{+w}{ }\PYG{n}{x}\PYG{+w}{ }\PYG{o+ow}{=}\PYG{+w}{ }\PYG{k+kt}{K}\PYG{+w}{ }\PYG{p}{(}\PYG{n+nf}{\PYGZbs{}}\PYG{n}{p}\PYG{+w}{ }\PYG{o+ow}{\PYGZhy{}\PYGZgt{}}\PYG{+w}{ }\PYG{n}{p}\PYG{+w}{ }\PYG{n}{x}\PYG{p}{)}
\PYG{+w}{    }\PYG{n}{fs}\PYG{+w}{ }\PYG{o}{\PYGZlt{}*\PYGZgt{}}\PYG{+w}{ }\PYG{n}{phis}\PYG{+w}{ }\PYG{o+ow}{=}\PYG{+w}{ }\PYG{k+kr}{do}
\PYG{+w}{        }\PYG{n}{f}\PYG{+w}{ }\PYG{o+ow}{\PYGZlt{}\PYGZhy{}}\PYG{+w}{ }\PYG{n}{fs}
\PYG{+w}{        }\PYG{n}{x}\PYG{+w}{ }\PYG{o+ow}{\PYGZlt{}\PYGZhy{}}\PYG{+w}{ }\PYG{n}{phis}
\PYG{+w}{        }\PYG{n}{return}\PYG{+w}{ }\PYG{p}{(}\PYG{n}{f}\PYG{+w}{ }\PYG{n}{x}\PYG{p}{)}

\PYG{k+kr}{instance}\PYG{+w}{ }\PYG{k+kt}{Monad}\PYG{+w}{ }\PYG{p}{(}\PYG{k+kt}{K}\PYG{+w}{ }\PYG{n}{r}\PYG{p}{)}\PYG{+w}{ }\PYG{k+kr}{where}
\PYG{+w}{    }\PYG{n}{return}\PYG{+w}{ }\PYG{o+ow}{=}\PYG{+w}{ }\PYG{n}{pure}
\PYG{+w}{    }\PYG{n}{phi}\PYG{+w}{ }\PYG{o}{\PYGZgt{}\PYGZgt{}=}\PYG{+w}{ }\PYG{n}{f}\PYG{+w}{ }\PYG{o+ow}{=}\PYG{+w}{ }\PYG{k+kt}{K}\PYG{+w}{ }\PYG{p}{(}\PYG{n+nf}{\PYGZbs{}}\PYG{n}{p}\PYG{+w}{ }\PYG{o+ow}{\PYGZhy{}\PYGZgt{}}\PYG{+w}{ }\PYG{n}{quantifier}\PYG{+w}{ }\PYG{n}{phi}\PYG{+w}{ }\PYG{p}{(}\PYG{n+nf}{\PYGZbs{}}\PYG{n}{x}\PYG{+w}{ }\PYG{o+ow}{\PYGZhy{}\PYGZgt{}}\PYG{+w}{ }\PYG{n}{quantifier}\PYG{+w}{ }\PYG{p}{(}\PYG{n}{f}\PYG{+w}{ }\PYG{n}{x}\PYG{p}{)}\PYG{+w}{ }\PYG{n}{p}\PYG{p}{))}

\PYG{c+c1}{\PYGZhy{}\PYGZhy{} sequence for quantifiers that depend on the moves played so far}
\PYG{n+nf}{hsequence}\PYG{+w}{ }\PYG{o+ow}{::}\PYG{+w}{ }\PYG{k+kt}{Monad}\PYG{+w}{ }\PYG{n}{m}\PYG{+w}{ }\PYG{o+ow}{=\PYGZgt{}}\PYG{+w}{ }\PYG{p}{[[}\PYG{n}{x}\PYG{p}{]}\PYG{+w}{ }\PYG{o+ow}{\PYGZhy{}\PYGZgt{}}\PYG{+w}{ }\PYG{n}{m}\PYG{+w}{ }\PYG{n}{x}\PYG{p}{]}\PYG{+w}{ }\PYG{o+ow}{\PYGZhy{}\PYGZgt{}}\PYG{+w}{ }\PYG{n}{m}\PYG{+w}{ }\PYG{p}{[}\PYG{n}{x}\PYG{p}{]}
\PYG{n+nf}{hsequence}\PYG{+w}{ }\PYG{o+ow}{=}\PYG{+w}{ }\PYG{n}{h}\PYG{+w}{ }\PYG{k+kt}{[]}
\PYG{+w}{  }\PYG{k+kr}{where}
\PYG{+w}{    }\PYG{n}{h}\PYG{+w}{ }\PYG{k+kr}{\PYGZus{}}\PYG{+w}{ }\PYG{k+kt}{[]}\PYG{+w}{       }\PYG{o+ow}{=}\PYG{+w}{ }\PYG{n}{return}\PYG{+w}{ }\PYG{k+kt}{[]}
\PYG{+w}{    }\PYG{n}{h}\PYG{+w}{ }\PYG{n}{s}\PYG{+w}{ }\PYG{p}{(}\PYG{n}{f}\PYG{+w}{ }\PYG{k+kt}{:}\PYG{+w}{ }\PYG{n}{fs}\PYG{p}{)}\PYG{+w}{ }\PYG{o+ow}{=}\PYG{+w}{ }\PYG{k+kr}{do}
\PYG{+w}{      }\PYG{n}{x}\PYG{+w}{  }\PYG{o+ow}{\PYGZlt{}\PYGZhy{}}\PYG{+w}{ }\PYG{n}{f}\PYG{+w}{ }\PYG{n}{s}
\PYG{+w}{      }\PYG{n}{xs}\PYG{+w}{ }\PYG{o+ow}{\PYGZlt{}\PYGZhy{}}\PYG{+w}{ }\PYG{n}{h}\PYG{+w}{ }\PYG{p}{(}\PYG{n}{s}\PYG{+w}{ }\PYG{o}{++}\PYG{+w}{ }\PYG{p}{[}\PYG{n}{x}\PYG{p}{])}\PYG{+w}{ }\PYG{n}{fs}
\PYG{+w}{      }\PYG{n}{return}\PYG{+w}{ }\PYG{p}{(}\PYG{n}{x}\PYG{+w}{ }\PYG{k+kt}{:}\PYG{+w}{ }\PYG{n}{xs}\PYG{p}{)}
\end{Verbatim}

\end{document}